\documentclass[]{fairmeta}
\usepackage{amsthm,amsmath,amssymb}
\usepackage{graphicx}
\usepackage{natbib}
\usepackage{subcaption}
\usepackage{booktabs}
\usepackage{algorithm}
\usepackage{algorithmic}
 \usepackage{multirow}

 \usepackage{footmisc}

\usepackage{makecell}

\newtheorem{theorem}{Theorem}[]

\newtheorem{remark1}[theorem]{Remark}

 \title{TeleBoost: A Systematic Alignment Framework for High-Fidelity, Controllable, and Robust Video Generation}
\author{TeleBoost Team}
\metadata[Corresponding Author]{Xuelong Li(\email{xuelong$\_$li@ieee.org})}

\begin{document}

\abstract{
Post-training is the decisive step for converting a pretrained video generator into a production-oriented model that is instruction-following, controllable, and robust over long temporal horizons. This report presents a systematical post-training framework that organizes supervised policy shaping, reward-driven reinforcement learning, and preference-based refinement into a single stability-constrained optimization stack. The framework is designed around practical video-generation constraints, including high rollout cost, temporally compounding failure modes, and feedback that is heterogeneous, uncertain, and often weakly discriminative. By treating optimization as a staged, diagnostic-driven process rather than a collection of isolated tricks, the report summarizes a cohesive recipe for improving perceptual fidelity, temporal coherence, and prompt adherence while preserving the controllability established at initialization. The resulting framework provides a clear blueprint for building scalable post-training pipelines that remain stable, extensible, and effective in real-world deployment settings.
}

\maketitle

\vspace{-0.1em}

\section{Introduction and Problem Setting}

Large-scale video diffusion and diffusion-transformer models have made rapid progress in recent years, extending generation from short, low-resolution clips to increasingly long, high-definition videos with complex motion and semantics~\citep{ho2020denoisingdiffusion,peebles2023scalable,chen2024videocrafter2,ma2024latte, yin2025slow}. Despite these advances, pretrained video generators remain far from satisfying real-world usage requirements~\citep{liu2024evalcrafter,wu2023tune}. In practice, their behavior is often sensitive to prompt phrasing, unstable over long temporal horizons, prone to localized artifacts (e.g., hands, text, fast motion), and weak at instruction following or controllable editing.

This gap between pretraining performance and deployment requirements motivates \emph{post-training}: a sequence of alignment and optimization procedures~\citep{liang2026integrating} applied after large-scale likelihood training. In contrast to pretraining, post-training operates under tight constraints on sampling cost, evaluator reliability, and system throughput, particularly in the video domain where rollouts are expensive and evaluation signals are noisy~\cite{xue2025dancegrpo,li2025selfpacedgrpo}.

\subsection{Why Video Post-Training Is Fundamentally Hard}

Post-training video generation models differs qualitatively from both language and image settings along several axes:

\paragraph{High rollout cost.}
A single video rollout involves dozens to hundreds of diffusion steps and frame decodes, often at high resolution~\citep{blattmann2023align,videodiffusionmodels}. This makes naive reinforcement learning or extensive trial-and-error optimization impractical~\citep{meng2025identity}.

\paragraph{Temporal structure and compounding errors.}
Video failures are rarely independent across frames. Localized artifacts, motion discontinuities, or identity drift can propagate over time~\citep{singer2022make,wu2023tune}, leading to compounding degradation that is not captured by frame-level metrics alone~\citep{xie2025progressive, yin2025slow,liang2023icocap}.

\paragraph{Ambiguous supervision.}
Text-to-video alignment is inherently many-to-many~\citep{he2024videoscore,liu2024evalcrafter}: a single prompt admits multiple valid renderings, and a generated video can often be described by several plausible captions. As a result, alignment rewards and preference judgments are intrinsically uncertain and weakly discriminative~\citep{liu2025videodpo,wu2025densedpo,liu2025bpgo}.

\paragraph{Evaluator brittleness.}
Most practical feedback signals—CLIP-like scorers~\cite{hessel2021clipscore,radford2021learning,wang2024videoclip}, VLM-based alignment models~\cite{liu2023visual,liu2025improving}, or handcrafted temporal metrics—are sensitive to decoding choices (fps, resolution, sampling strategy) and can saturate or bias optimization as the generator improves~\citep{li2025selfpacedgrpo,meng2025identity}.

These properties imply that post-training video models is not simply a matter of ``adding RL'', but requires careful system-level design to manage uncertainty, structure learning signals, and maintain stability throughout optimization.

\subsection{Post-Training as a Staged Optimization Pipeline}

A central premise of this report is that effective video post-training is most naturally organized as a staged optimization process. In practice, different alignment objectives become meaningful at different points in training, and attempting to optimize all goals simultaneously often leads to instability, weak supervision, or reward exploitation~\citep{xue2025dancegrpo}.

We therefore organize post-training into three sequential stages, each with a distinct role:

\begin{enumerate}
    \item \textbf{Stage I: Supervised Fine-Tuning (SFT).}  
    SFT~\citep{ouyang2022training,tunstall2023zephyr} adapts the pretrained backbone to follow instructions and controllable interfaces, reduces catastrophic failure modes, and establishes a stable reference policy for downstream optimization.

    \item \textbf{Stage II: Reinforcement Learning via GRPO.}  
    Group-based reinforcement learning~\citep{shao2024deepseekmath,xue2025dancegrpo,meng2025identity,park2025deepvideo} optimizes measurable objectives—such as perceptual quality and temporal coherence—using relative comparisons within prompt groups, while avoiding unstable value critics.

    \item \textbf{Stage III: Direct Preference Optimization (DPO).}  
    Preference-based optimization~\citep{rafailov2023direct,liu2025videodpo,cheng2025discriminator,jiang2025huvidpo,wu2025densedpo} incorporates holistic human judgments that are difficult to encode as explicit rewards, serving as a final alignment step once low-level quality and stability have been established.
\end{enumerate}

% ---------- Figure: Systematic Overview (place near the end of
% \subsection{Post-Training as a Staged Optimization Pipeline} ----------
\begin{figure*}[t]
    \centering
    \includegraphics[width=\linewidth]{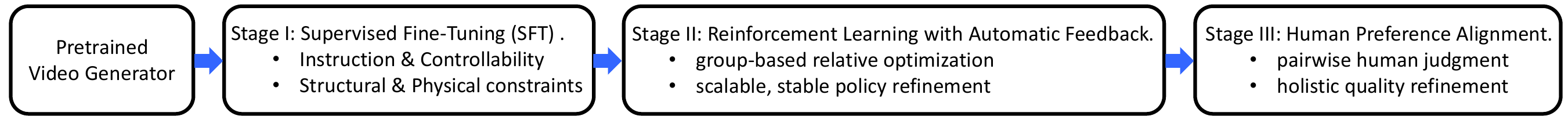}
    \caption{
    Systematic overview of our video post-training framework.
    Starting from a pretrained video diffusion backbone, the pipeline proceeds through three staged optimizations:
    (I) supervised adaptation to establish a stable and controllable policy,
    (II) automatic-feedback optimization to improve alignment, perceptual quality, and temporal coherence under stability constraints, and
    (III) human-preference alignment to capture holistic judgments that are difficult to encode as explicit rewards.
    Evaluation and diagnostics act as cross-stage components, enabling slice-based analysis and reproducible validation.
    }
    \label{fig:posttrain_overview}
\end{figure*}

As summarized in Fig.~\ref{fig:posttrain_overview}, we organize video post-training into a staged optimization. This staged design reflects both practical constraints and empirical observations: early stages shape the feasible behavior space, mid-stage optimization improves measurable properties under controlled feedback, and late-stage preference learning refines global perceptual and semantic quality.

\subsection{Design Principles Guiding the Pipeline}

Across all stages, the pipeline is guided by three recurring design principles:

\paragraph{Reliability of feedback.}
Optimization should account for uncertainty and bias in evaluators~\citep{gao2023scaling,lambert2025rewardbench} rather than assuming rewards are uniformly trustworthy~\citep{liu2025bpgo}. This is especially critical for ambiguous prompts and late-stage training.

\paragraph{Structural alignment of learning signals.}
Learning pressure should be applied where errors occur—in space, time, or semantic structure—rather than uniformly across all model components~\citep{wu2025densedpo}.

\paragraph{Adaptivity over training time.}
As the generator improves, feedback signals may saturate or lose discriminative power~\cite{liang2022simple,liang2022penalizing}. Post-training mechanisms must therefore adapt supervision strategies as competence increases.

These principles inform the design of the reinforcement learning stage and its extensions, as well as the role of preference optimization as a complementary signal.

\subsection{Scope and Organization of This Report}

This report presents a systematic post-training framework for video diffusion models built around the staged pipeline above. The emphasis is not on proposing a single algorithm, but on describing a cohesive methodology that integrates data design, optimization, feedback modeling, and system infrastructure.

The remainder of the report is organized as follows:
\begin{itemize}
    \item Chapter~2 introduces the overall post-training pipeline, defining the inputs, outputs, and responsibilities of each stage.
    \item Chapter~3 details Stage~I (SFT), including standard instruction tuning and joint supervision with physical priors.
    \item Chapter~4 presents Stage~II (GRPO-based reinforcement learning) and its extensions for reliability, structure, and adaptivity.
    \item Chapter~5 focuses on Stage~III (DPO), with emphasis on diffusion-specific formulations and video preference data construction.
    \item Chapters~6--8 cover reward modeling, infrastructure, and experimental validation.
\end{itemize}

Together, these chapters aim to present video post-training not as a collection of isolated techniques, but as a coherent, reproducible optimization pipeline designed for real-world generative systems.

\section{Stage I: Supervised Fine-Tuning (SFT)}

In our post-training system, supervised fine-tuning (SFT) is designed as a deliberate and essential policy-shaping stage rather than a lightweight adaptation step~\citep{wei2021finetuned,ouyang2022training,touvron2023llama}. Starting from a large-scale pretrained video diffusion backbone, we use SFT to explicitly constrain the generator’s behavior space, stabilize generation dynamics, and construct a reliable reference policy for all subsequent optimization stages.

Throughout development, we found that downstream reinforcement learning and preference optimization are highly sensitive to the quality and structure of the initial supervised policy. As a result, we invest substantial effort in SFT, treating it as a first-class system component rather than a preparatory procedure.

\subsection{Design Principles of SFT in Our System}

Our SFT design follows three guiding principles derived from extensive empirical iteration:

\begin{itemize}
    \item \textbf{Policy shaping over quality optimization.}  
    SFT is used to define what the model is allowed to do, not to exhaustively optimize how good the outputs look.

    \item \textbf{Stability before expressiveness.}  
    We prioritize rollout stability, structural coherence, and controllability, as unstable policies significantly degrade the effectiveness of later post-training stages.

    \item \textbf{Unified supervision with modular constraints.}  
    All supervised signals are integrated into a single training framework, with different constraints targeting complementary aspects of model behavior.
\end{itemize}

Under these principles, SFT serves as the structural foundation of the entire post-training pipeline.

\begin{figure*}[t]
    \centering
    \includegraphics[width=0.35\linewidth]{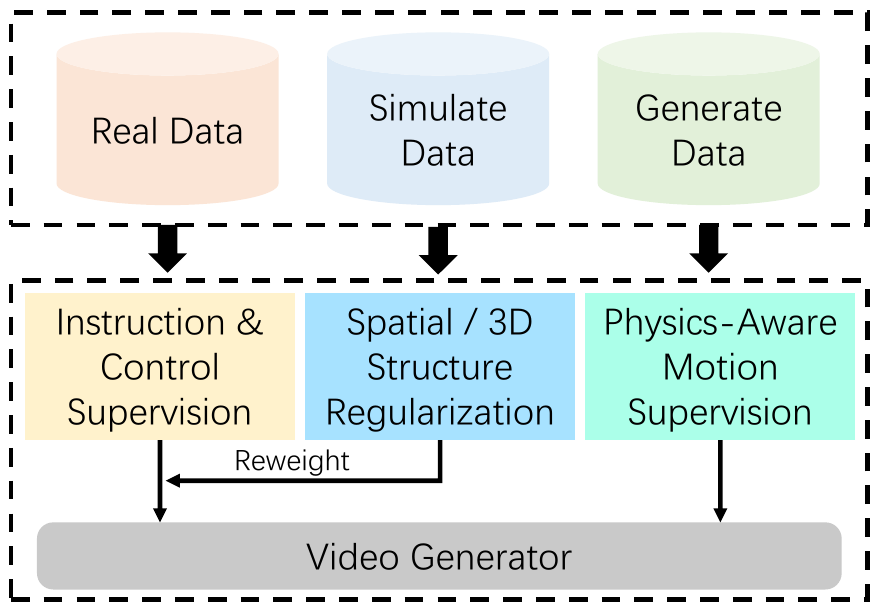}
    \caption{
    Overview of Stage~I supervised fine-tuning (SFT).
    Starting from a pretrained video diffusion backbone with a frozen encoder and diffusion
    transformer, SFT shapes the generator through a unified training objective that integrates
    instruction and control supervision, spatial-structure–aware constraints, and physics-aware
    motion supervision.
    All structural priors are injected conservatively at the decoder level, producing a stable and
    structurally constrained reference policy for subsequent post-training stages.
    }
    \label{fig:sft_framework}
\end{figure*}

\subsection{Instruction and Control-Oriented SFT}

We begin SFT by aligning the pretrained video diffusion model with user-facing instruction patterns and controllable generation interfaces~\citep{yang2024cogvideox,blattmann2023align,chen2024videocrafter2}. Instead of relying solely on generic text--video supervision, we curate instruction-centric training data that more closely reflects real-world usage scenarios~\citep{liang2019vrr,chen2024hydra} encountered during deployment and evaluation.

This baseline supervision emphasizes:
\begin{itemize}
    \item explicit descriptions of temporal structure and event ordering,
    \item camera behavior and viewpoint changes (e.g., rotation, translation, dolly)~\citep{zheng2024cami2v,he2024cameractrl},
    \item compositional constraints and editing-oriented instructions.
\end{itemize}

Although instruction- and control-oriented samples constitute only a minority of the overall SFT data, they play a critical role in stabilizing rollouts and reducing ambiguity during downstream evaluation. In practice, this supervision establishes a baseline policy that is predictable, compositional, and responsive to structured prompts, providing a reliable interface for subsequent post-training stages.

However, we observe that instruction and control supervision alone is insufficient to address deeper spatial and geometric failure modes, particularly under large camera motions or long temporal horizons~\citep{xing2024dynamicrafter}. This motivates the introduction of structure-aware supervision mechanisms described in the following sections.

\subsection{Spatial-Structure–Aware SFT with 3D Consistency Constraints}

A consistent limitation observed in large-scale video diffusion models is the degradation of three-dimensional structure under camera motion~\citep{ren2025gen3c,kuang2024collaborative}. Even with strong instruction following and controllability, generated videos may exhibit background collapse, object deformation, or unstable depth ordering as viewpoint changes accumulate over time. These artifacts are particularly common under explicit camera rotations, translations, or orbiting trajectories.

From a post-training perspective, such geometric instabilities are undesirable not only because they affect perceptual quality, but also because they are weakly constrained by standard alignment and aesthetic evaluators. As a result, these errors often persist into later optimization stages, where they introduce high-variance feedback and complicate reinforcement learning.

\paragraph{Structure-sensitive supervision.}
To address this issue early, we incorporate a spatial-structure–aware supervision component into the SFT stage. The goal is not to recover explicit 3D geometry, but to bias the supervised policy toward structurally stable generation under viewpoint change, thereby reducing downstream optimization burden.

Specifically, we employ a structural assessment signal that captures scene-level stability and object-level geometric consistency across frames. This signal is derived automatically from a mixture of real-world videos, simulator-generated data, and model-generated sequences covering diverse camera motions, without requiring manual annotation.

\paragraph{Loss reweighting for geometric stability.}
The structural signal is integrated through loss reweighting during SFT. Samples exhibiting geometric distortion or spatial collapse receive stronger corrective gradients, while structurally stable samples are deemphasized. This mechanism preserves the original diffusion objective while concentrating learning capacity on failure cases that are insufficiently constrained by standard supervised losses.

In practice, this spatially informed SFT significantly improves geometric stability under camera motion and reduces structurally induced noise in downstream reward signals. As a result, subsequent reinforcement learning operates on more stable rollouts and exhibits improved optimization efficiency for objectives related to temporal coherence and global scene consistency.

\subsection{Physics-Aware Supervised Fine-Tuning}

Another prominent source of instability arises from violations of physical dynamics, particularly in scenarios involving fluids and deformable materials~\citep{ummenhofer2019lagrangian,sanchez2020learning,li2018learning}. We find that such errors are weakly penalized by most downstream evaluators and therefore persist if not addressed explicitly.
We incorporate physics-aware supervision into SFT to bias the generator toward physically plausible motion patterns from the outset.

\paragraph{Joint real-and-simulated training.}
We construct a joint training setup that combines real-world fluid videos with physics-based simulation data~\citep{raissi2019physics,karniadakis2021physics,jin2021nsfnets}. Simulation data provides precise motion supervision under controlled conditions, while real-world videos prevent overfitting to synthetic dynamics.

\paragraph{Auxiliary motion modeling.}
On top of the standard RGB decoding pathway, we introduce an auxiliary motion prediction branch trained to estimate inter-frame motion. The motion branch is supervised using ground-truth optical flow~\citep{teed2020raft,sun2018pwc} extracted from training data, and its intermediate features are fused into the RGB decoder via a lightweight, zero-initialized module~\citep{shi2024motion,hu2022lora,zhang2023adding}.

Throughout this process, the pretrained backbone remains frozen and only decoder-level parameters are updated.

\paragraph{Effect on downstream optimization.}
By improving physical plausibility at the supervised stage, we significantly reduce the variance and ambiguity of reward signals for physics-related prompts, enabling more effective reinforcement learning and preference optimization.

\subsection{SFT as the Foundation for Reinforcement Learning}

In our system, supervised fine-tuning is not intended to fully resolve alignment or subjective quality optimization. Instead, it deliberately establishes a well-behaved and structurally constrained reference policy that serves as the foundation for all subsequent post-training stages~\citep{ouyang2022training,bai2022constitutional,christiano2017deep}.

By addressing major sources of instability—such as inconsistent instruction following, spatial drift, and physically implausible motion—during SFT, we ensure that downstream reinforcement learning operates on stable rollouts and interpretable feedback signals. This substantially improves the effectiveness of group-relative objectives and preference-based optimization, allowing later stages to focus on relative quality, preference modeling, and long-horizon behavior refinement rather than correcting fundamental structural deficiencies.

\section{Stage II: Reinforcement Learning via GRPO}

Stage~II constitutes the central optimization phase of our post-training pipeline. Starting from the stable reference policy $\pi_{\mathrm{ref}}$ established in Stage~I (SFT), we employ Reinforcement Learning (RL) to drive measurable improvements in semantic alignment, perceptual fidelity, and temporal coherence using automatically computable feedback~\citep{liang2026integrating}. Unlike SFT, which primarily constrains the feasible behavior space via static supervision~\citep{ouyang2022training}, Stage~II is fundamentally defined by a closed-loop process of \emph{relative optimization}~\citep{schulman2017ppo}: it refines model behavior by exploiting comparative signals among generated samples (\emph{sample $\rightarrow$ evaluate $\rightarrow$ update}).

From a system perspective, reinforcement learning for video generation operates under a set of rigorous practical constraints. Video rollouts are high-dimensional and computationally expensive~\citep{videodiffusionmodels,blattmann2023align}; feedback signals are diverse, noisy, and often imperfect; and stability with respect to the SFT reference policy must be strictly preserved to prevent catastrophic forgetting~\citep{ramasesh2021effect}. These factors do not define a single algorithmic bottleneck but instead shape how optimization objectives should be structured, combined, and scheduled.

Accordingly, we implement Stage~II not as a single monolithic objective, but as a unified optimization stack built on a stable group-relative backbone (GRPO) augmented by modular refinement mechanisms. This design allows us to address the unique, endogenous challenges of video generation systematically:

\begin{itemize}
    \item \textbf{The Spatiotemporal Credit Assignment Problem:} Video generation failures are rarely uniform. A single scalar reward (e.g., ``score 0.7") is excessively coarse, failing to inform the model \emph{where} (spatial location) or \emph{when} (temporal segment) a failure occurred~\citep{mesnard2020counterfactual}—such as momentary hand deformation or transient background flickering. This results in sparse, inefficient optimization signals. To address this, we introduce {ViPO (Visual Preference Policy Optimization)}, which lifts scalar feedback into structured spatiotemporal advantage maps.
    \item \textbf{Ambiguity and Trust Allocation:} Visual generation mappings—whether Text-to-Video (T2V) or Image-to-Video (I2V)—are inherently many-to-many~\citep{singer2022make,he2024videoscore}. A single text prompt or starting image can validly lead to diverse outputs. Existing reward models often exhibit inconsistent variance across these ambiguous mappings~\citep{gao2023scaling,lambert2025rewardbench}. Treating feedback from all prompt groups with equal trust risks overfitting to evaluator noise. We address this via {BPGO (Bayesian Prior-Guided Optimization)}, which utilizes Bayesian priors to dynamically allocate optimization ``trust" and filter unreliable signals.
    \item \textbf{Non-stationarity and Reward Saturation:} As the generator's capabilities improve, static reward models frequently saturate or lose discriminative power~\citep{skalse2022defining}, leading to vanishing gradients. We design {Self-Paced GRPO} to construct an adaptive reward curriculum that co-evolves with the generator's competence, preventing late-stage training stagnation.
    \item \textbf{Multi-Objective Trade-offs:} Production-grade video generation demands balancing conflicting dimensions such as image quality, motion smoothness, semantic alignment, and safety~\citep{wu2023fine}. We employ a {Joint Reward} mechanism to dynamically manage these trade-offs within a unified loss function.
\end{itemize}

This chapter details this systematic GRPO technology stack.

\begin{figure*}[t]
    \centering
    \includegraphics[width=\textwidth]{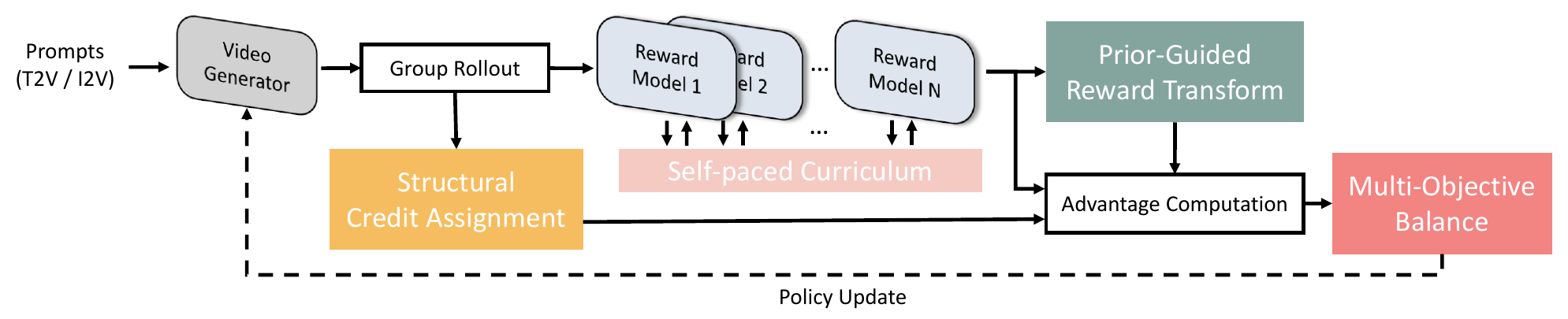}
    \caption{Stage-II GRPO optimization stack.
    Starting from the SFT-initialized policy, Stage~II follows a closed-loop \emph{sample $\rightarrow$ evaluate $\rightarrow$ update} pipeline.
    For each prompt group, the current policy generates multiple video rollouts, which are scored by a set of evaluators (e.g., visual quality, motion/temporal coherence, text alignment, safety, etc.).
    GRPO performs group-relative normalization to convert raw feedback into stable learning signals.
    On top of this backbone, we introduce modular refinements: structural credit Assignment (ViPO) lifts scalar feedback into spatiotemporal advantage maps for fine-grained credit assignment; prior-guided reward transform (BPGO) calibrates the trust of noisy/ambiguous supervision via prior-referenced uncertainty; and self-paced curriculum (Self-Paced GRPO) constructs an adaptive reward curriculum to mitigate reward saturation and late-stage stagnation.
    A multi-objective balance (Joint Reward) layer reconciles multi-objective trade-offs and produces the final optimization signal used to update the policy while maintaining stability.}
    \label{fig:grpo_framework}
\end{figure*}

\subsection{GRPO Baseline: The Critic-Free Backbone}

As the foundation for all subsequent methods, we adopt {Group Relative Policy Optimization (GRPO)}~\citep{shao2024deepseekmath}. In contrast to methods relying on value function estimation (e.g., PPO~\citep{schulman2017ppo}), GRPO eliminates the dependency on a critic network by leveraging group-relative advantages. This is particularly critical for high-dimensional video generation, where training a value critic that remains accurate and stable under distribution shifts is computationally prohibitive.

The standard GRPO protocol operates as follows:
For a given prompt $c$, the policy $\pi_\theta$ samples a group of videos $\{v_1, \dots, v_G\}$. Each video $v_i$ receives a scalar score $r_i$ from the reward function. To reduce variance, GRPO computes group-relative advantages $A_i$ by standardizing scores within the group:
\begin{equation}
    A_i = \frac{r_i - \text{mean}(\{r_1, \dots, r_G\})}{\text{std}(\{r_1, \dots, r_G\}) + \epsilon}
\end{equation}
The policy is then updated by maximizing a surrogate objective weighted by these advantages, subject to a KL divergence constraint~\citep{ziegler2019fine} (typically estimated via a per-token KL penalty) to prevent deviation from the reference policy $\pi_{\mathrm{ref}}$. This ``within-group comparison" mechanism is naturally suited for video tasks, as it transforms the absolute scoring regression problem into a robust relative ranking problem, making the optimization invariant to the absolute scale of the reward model.

\subsection{ViPO: Seeing What Matters via Structural Credit Assignment}

Conventional GRPO treats a video as a monolithic entity, assigning it a single scalar advantage. However, visual defects are often localized in space and time. For instance, a video might be cinematically perfect but exhibit a hand collapse at the 3-second mark. If assigned a low global scalar score, the model receives no gradient information indicating that the ``hand" was the issue; conversely, it may erroneously suppress well-generated elements like background or lighting. This coarse-grained feedback leads to inefficient credit assignment, as the model struggles to disentangle local failures from global quality.

\begin{figure}
	\centering
	\includegraphics[width=0.6\linewidth]{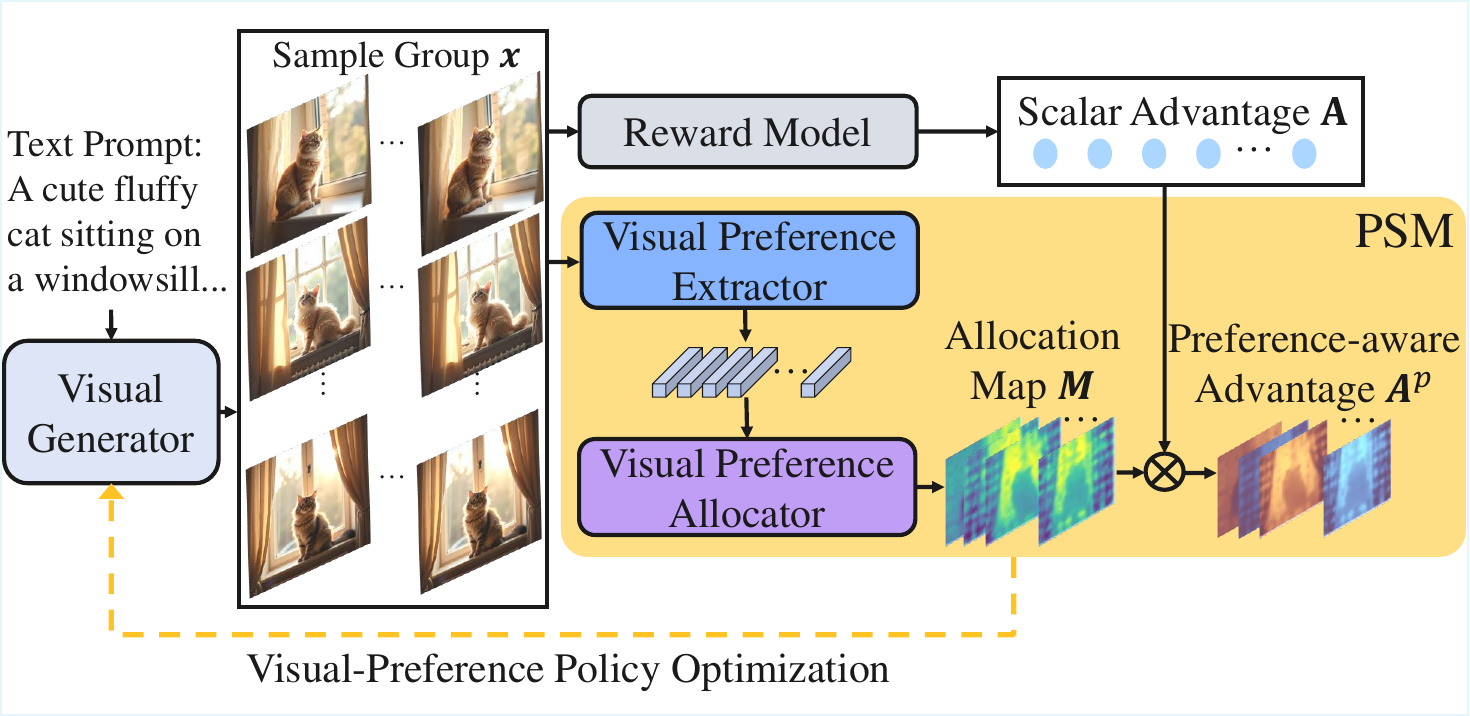}
	\caption{ViPO: a perceptual structuring module builds spatial/temporal allocation maps used to convert scalar GRPO advantages into pixel/latent-level advantages.}
	\label{fig:framework-vipo}
\end{figure}

To address this, we introduce \textbf{Visual Preference Policy Optimization (ViPO)} \citep{ni2025vipo}, which ``back-projects" scalar advantages into pixel or latent space without requiring dense supervision. Specifically, ViPO introduces a \textbf{Perceptual Structuring Module (PSM)} that employs frozen, pretrained visual backbones (e.g., DINOv2~\citep{oquab2023dinov2} or VideoMAE~\citep{tong2022videomae}) to extract spatiotemporal feature maps from generated videos. By analyzing feature saliency or alignment, ViPO constructs a fine-grained \textbf{Advantage Map} $M \in \mathbb{R}^{T \times H \times W}$. The optimization objective is then reformulated to weight the policy gradient at each spatiotemporal position by its corresponding local advantage:
\begin{equation}
    \mathcal{L}_{\text{ViPO}} = \mathbb{E} \left[ \sum_{t,h,w} M_{i}^{(t,h,w)} \cdot A_i \cdot \log \pi_\theta(v_i) \right]
\end{equation}
This mechanism effectively redirects gradients toward the most visually critical or failure-prone regions. By structurally decomposing the learning signal, ViPO enables the model to perform precise correction of localized artifacts (e.g., fixing a specific object's motion) while preserving the global structure, significantly enhancing sample efficiency.

\subsection{BPGO: Learning What to Trust under Ambiguity}

Visual generation tasks—whether {Text-to-Video (T2V)} or {Image-to-Video (I2V)}—face inherent {``many-to-many" ambiguity}~\citep{chen2025seed}. In T2V, a single text prompt corresponds to infinite plausible visual realizations. In I2V, a static starting image can validly evolve into diverse motion trajectories. Reward models often exhibit inconsistent confidence and noise across these complex variations—confident in unambiguous cases but noisy in others. Directly optimizing these raw rewards can cause the model to overfit to evaluator noise in ambiguous regions. A mechanism is required to distinguish reliable feedback from noise across both tasks.

\begin{figure}
	\centering
	\includegraphics[width=\linewidth]{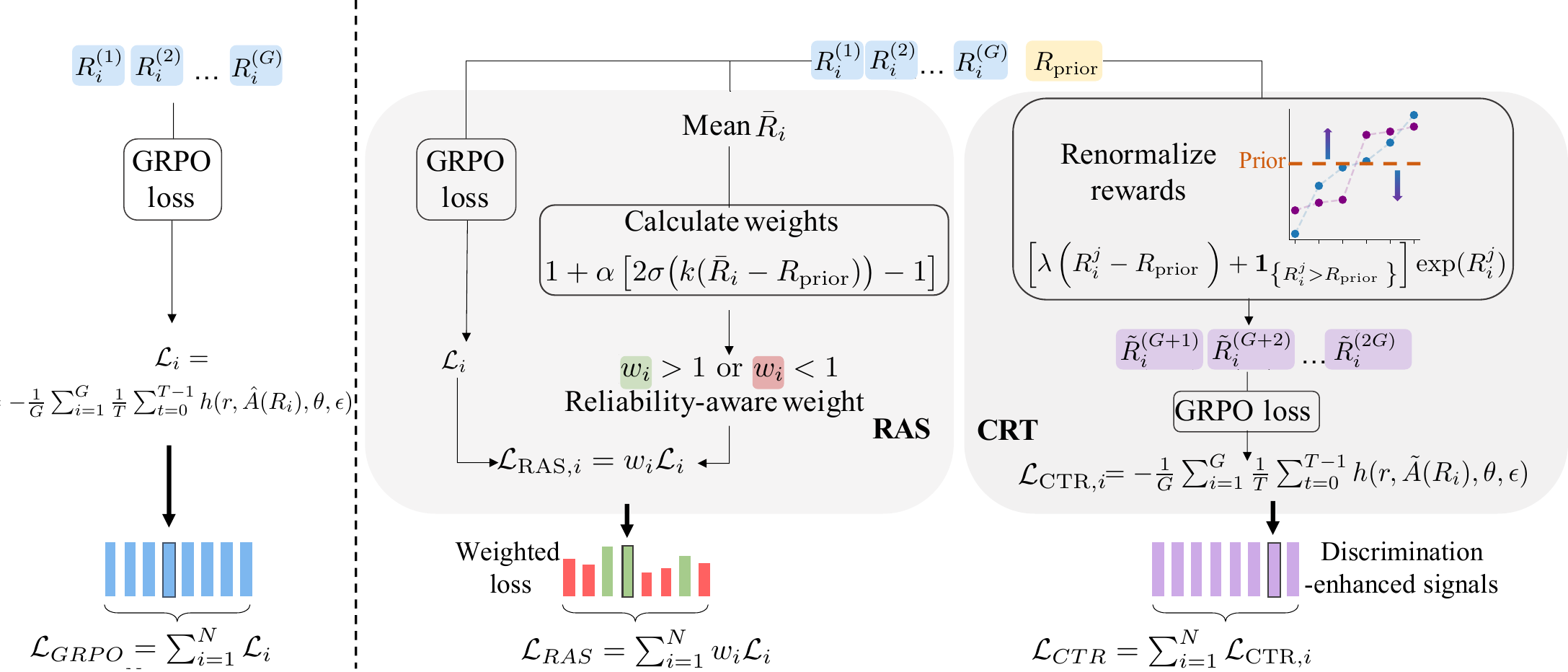}
	\caption{BPGO: Reliability-Adaptive Scaling (RAS) reweights prompt groups based on deviation from prior rewards; Contrastive Reward Transformation (CRT) sharpens within-group discrimination.}
	\label{fig:framework-bpgo}
\end{figure}

\textbf{Bayesian Prior-Guided Optimization (BPGO)} \citep{liu2025bpgo} solves this by utilizing a {Bayesian Prior} to guide the optimization process. The core concept is to employ a prior distribution (typically derived from the SFT model's historical performance or reference statistics) as a ``Trust Anchor." BPGO incorporates two hierarchical mechanisms:
\begin{enumerate}
    \item \textbf{Inter-group Trust Allocation (Reliability-Adaptive Scaling):} The system compares the reward distribution of the current rollout group against the prior. If a group's distribution deviates significantly from the prior with high variance, it is flagged as ``ambiguous" or ``unreliable," and its optimization weight is dynamically down-scaled. Conversely, groups with clear, low-variance signals are up-weighted.
    \item \textbf{Intra-group Prior-anchored Renormalization (Contrastive Reward Transformation):} When computing advantages, BPGO considers the prior score as a baseline. It stretches the advantage values of high-confidence positive samples (those significantly beating the prior) while compressing the advantage space of ambiguous samples.
\end{enumerate}
Through this approach, BPGO enables the model to ``Learn What to Trust", preventing reward hacking on noisy prompts while aggressively optimizing on high-confidence signals, effectively supporting both T2V and I2V modalities.

\subsection{Self-Paced GRPO: Growing with the Generator}

In large-scale video post-training, we observe a significant phenomenon: {Reward Saturation}. In early training stages, reward models based on coarse quality or simple semantics provide effective guidance. However, as the generator improves, the quality of generated videos rapidly surpasses the discrimination threshold of these static models, causing scores to cluster near the ceiling with vanishing variance. Consequently, the intra-group advantage $A_i$ collapses to noise, causing training to stagnate.

\textbf{Self-Paced GRPO} \citep{li2025selfpacedgrpo} addresses this via a {competence-aware} dynamic curriculum framework. Instead of using a fixed, static reward function, this method treats reward signals as an evolving curriculum that adapts to the generator's current stage. The system monitors the generator's performance statistics (e.g., reward distribution sparsity, intra-group discriminability) in real-time to transition between phases:
\begin{itemize}
    \item \textbf{Phase 1 (Visual Fidelity):} Focuses on basic visual quality and dynamic fluency. The system prioritizes rewards that penalize obvious structural collapse.
    \item \textbf{Phase 2 (Temporal Coherence):} As basic metrics saturate (indicated by low variance), the system automatically shifts weight to temporal consistency and complex motion logic rewards.
    \item \textbf{Phase 3 (Semantic Alignment):} The final phase focuses on fine-grained text alignment and aesthetic details, which are only optimizing when the structural foundations are stable.
\end{itemize}
This ``Growing with the Generator" strategy ensures that the model always faces a ``zone of proximal development,"~\citep{bengio2009curriculum} receiving challenging yet informative feedback signals throughout the entire training lifecycle.

\subsection{Joint Reward: Multi-Objective Balancing}

% \emph{[This section is reserved for the detailed implementation of the multi-objective joint reward system. It will discuss the specific composition of reward models (e.g., combining Quality, Motion, and Alignment scores), dynamic weighting strategies based on gradient norms or Pareto-frontier analysis, and conflict resolution mechanisms to handle competing objectives, such as balancing aesthetic style against strict prompt adherence.]}
Video generation post-training naturally induces a multi-objective optimization problem, where multiple reward models capture different dimensions of human preferences (e.g., quality, motion, and text–video alignment).
While these rewards provide complementary signals, simple advantage summation often yields conflicting update directions, manifesting as imbalanced progress across objectives and indicating complex inter-reward coupling.

We therefore cast joint reward optimization as a multi-objective optimization problem~\citep{yu2020gradient}, aiming to balance competing objectives rather than optimize them independently. Conventional approaches operate at the gradient level, seeking a conflict-aware update direction by solving a minimum-norm problem:
\begin{equation}
\min_{\{c_i\}_{i=1}^{N}}\Bigl\|\sum_{i=1}^{N}c_i\nabla_{\!\theta}\mathcal{L}_i(\theta)\Bigr\|^{2}\quad\mathrm{s.t.}\quad \sum_ic_i=1,\;\;c_i\geq 0\;\;\forall i,
\end{equation}
where $c_i$ denotes the weight of the $i$-th reward and $\nabla_{\!\theta}\mathcal{L}_i(\theta)$ its corresponding gradient. While effective at mitigating gradient conflicts, such methods incur prohibitive memory and computation costs for large-scale video generation models.

To improve scalability, we reformulate the objective at the advantage level, avoiding explicit gradient computation:
\begin{equation}
\min_{\{c_i\}_{i=1}^{N}}\Bigl\|\sum_{i=1}^{N}c_iA_i\Bigr\|^{2}\quad\mathrm{s.t.}\quad \sum_ic_i=1,\;\;c_i\geq 0\;\;\forall i,
\end{equation}
where $A_i$ denotes the advantage induced by the $i$-th reward model. The final optimization objective is computed as:
\begin{equation}
    \mathcal{L} = \sum_i c_i\mathcal{L}_i.
\end{equation}
This formulation retains the benefits of multi-objective balancing while substantially reducing memory overhead, enabling stable and efficient joint reward optimization in large-scale video post-training.

\subsection{Summary}

By integrating these modular advancements, our Stage~II transcends a simple PPO loop to become a perceptually aware and adaptive optimization system. ViPO addresses the spatial question of ``Where to learn," BPGO addresses the reliability question of ``What to trust," and Self-Paced GRPO addresses the temporal question of ``When to learn." These three components, organically integrated with the critic-free GRPO backbone, collectively underpin a high-performance, stable, and scalable post-training workflow for video generation.

\section{Stage III: Direct Preference Optimization (DPO)}

The final stratum of our post-training stack employs Direct Preference Optimization (DPO)~\citep{rafailov2023direct} to align the generator with holistic human preferences that resist quantification by scalar reward models. While Stage~II (GRPO) effectively drives the policy toward measurable objectives—such as temporal consistency and prompt adherence—it remains bounded by the expressiveness of the reward function and the exploration efficiency of online RL~\citep{ouyang2022training,christiano2017deep}. Stage~III serves as a targeted \emph{offline alignment} phase, refining the policy using curated pairs of preferred and dispreferred trajectories to capture subjective qualities like cinematic pacing, aesthetic composition, and nuanced motion realism.

Although methodologically distinct from the online rollouts of GRPO, we frame DPO within the same reinforcement learning paradigm: it optimizes the policy to maximize an implicit reward function defined by the preference data, subject to a KL divergence constraint~\citep{schulman2017ppo,ziegler2019fine} against the reference model. In our pipeline, this stage is critical for the ``last mile" of alignment, correcting distribution shifts and mode-seeking behaviors introduced during the aggressive optimization of Stage~II.

\subsection{DPO for Diffusion-Based Video Generation}

We adapt the DPO objective, originally formulated for autoregressive models, to the continuous diffusion setting. Following recent advancements in diffusion alignment \citep{rafailov2023direct}, we express the loss in terms of the evidence lower bound (ELBO) or direct denoising error surrogates.

Formally, given a dataset of triplets $\mathcal{D} = \{x, y_w, y_l\}$ containing a prompt $x$, a preferred video $y_w$, and a dispreferred video $y_l$, the objective minimizes:
\begin{equation}
    \mathcal{L}_{\mathrm{DPO}}(\pi_\theta; \pi_{\mathrm{ref}}) = -\mathbb{E}_{(x, y_w, y_l) \sim \mathcal{D}} \left[ \log \sigma \left( \beta \log \frac{\pi_\theta(y_w|x)}{\pi_{\mathrm{ref}}(y_w|x)} - \beta \log \frac{\pi_\theta(y_l|x)}{\pi_{\mathrm{ref}}(y_l|x)} \right) \right]
\end{equation}
where $\pi_{\mathrm{ref}}$ is the frozen checkpoint from Stage~II. Unlike standard SFT which maximizes the likelihood of $y_w$ in isolation, this formulation explicitly suppresses the likelihood of $y_l$ relative to $y_w$, forcing the model to unlearn specific failure modes (e.g., ``zombie" motion, static frames, or warped geometries) that SFT alone cannot easily penalize.

\subsection{Preference Data Construction Strategy}

The efficacy of Stage~III is determined less by the loss formulation and more by the rigorous construction of the preference dataset $\mathcal{D}$. Unlike language modeling, where annotators compare static text, video preference annotation faces challenges of temporal dimensionality and presentation bias~\citep{wu2023fine}. We employ a three-pronged strategy to construct high-signal preference pairs:

\paragraph{1. Policy-On-Policy Hard Negatives.}
To correct the specific pathologies of the Stage~II model, we generate pairs directly from the current policy $\pi_{\mathrm{GRPO}}$. We employ an ensemble of VLM-based critics (e.g., Video-LLaVA~\citep{lin2024video}, Gemini-Vision) and heuristic filters (optical flow magnitude, aesthetic scorers~\citep{schuhmann2022laion}) to automatically rank these samples. Pairs with high semantic overlap but distinct visual quality (e.g., one with valid motion, one with object collapse) are selected as high-information training signals. This ``self-correction" loop ensures the model learns to avoid its own most probable errors.

\paragraph{2. Synthetic Temporal Negatives (Discriminator-Free Construction).}
Addressing the scarcity of high-quality pairwise video data, we adopt a discriminator-free approach for temporal alignment \citep{cheng2025discriminator}. We treat high-quality real videos (or successful generations)~\citep{videodiffusionmodels,singer2022make} as $y_w$. For $y_l$, we procedurally generate ``hard temporal negatives" by perturbing $y_w$:
\begin{itemize}
    \item \textbf{Temporal Reversal:} Reversing the frame order to penalize unnatural physical dynamics (e.g., water flowing up, smoke condensing).
    \item \textbf{Frame Shuffling:} Randomizing local frame order to enforce causality and smoothness.
    \item \textbf{Stalling/Freezing:} Duplicating frames to penalize static video generation under motion-heavy prompts.
\end{itemize}
Optimizing against these synthetic negatives forces the model to learn a strictly causal and temporally coherent probability density, which is often under-constrained by image-based pretraining.

\paragraph{3. Holistic Human-Aligned Ranking.}
For the highest tier of alignment—cinematic quality and style—we utilize a smaller, high-quality dataset annotated by human experts. These annotations focus on ``soft" criteria that metrics~\cite{he2024videoscore} fail to capture, such as lighting consistency, narrative logicality, and emotional tone. We utilize these pairs to fine-tune the reward scale $\beta$, preventing the over-optimization of metrics at the expense of visual naturalness.

\subsection{System Integration and Training Protocol}

\paragraph{Reference Policy Management.}
Strict adherence to the reference policy $\pi_{\mathrm{ref}}$ is paramount. We initialize $\pi_{\mathrm{ref}}$ from the final Stage~II checkpoint rather than the SFT model. This ensures that DPO refines the RL-optimized frontier rather than reverting to the mean SFT behavior. We observe that relaxing the KL constraint (low $\beta$) leads to visual degeneration, while excessive constraint (high $\beta$) prevents the correction of established artifacts. We employ a dynamic $\beta$ schedule, starting high to preserve stability and annealing slightly to allow exploration of the preference frontier.

\paragraph{Resolution and Compute Efficiency.}
Due to the memory overhead of storing gradients for two denoising paths (winner and loser), DPO training is computationally intensive. We adopt a mixed-resolution strategy: preference gradients are primarily computed at lower resolutions or shorter durations (e.g., 480p, 2s) where structural failures are most apparent, while the visual encoder features are often cached or frozen to reduce memory footprint. This allows us to scale batch sizes sufficient for stable contrastive learning.

\subsection{Summary}
Stage~III completes the post-training triad by explicitly aligning the generation probability mass with human-preferred outcomes. By systematically suppressing the likelihood of temporal artifacts and policy-specific failures through DPO, we achieve a final model that is not only metric-compliant (via GRPO) but also aesthetically superior and robust to the nuanced failure modes inherent in video diffusion.

\section{Reward Modeling and Signal Orchestration}
\label{sec:reward-modeling}

In our post-training framework, reward modeling is approached not merely as the training of scoring functions, but as the architectural design of a \textbf{heterogeneous feedback system}~\citep{bai2022constitutional,touvron2023llama}. Video optimization presents a high-dimensional landscape where the efficacy of algorithms like GRPO is fundamentally bounded by the discriminative power and reliability of the guiding signal~\citep{gao2023scaling,coste2023reward}. A saturating or noisy reward signal inevitably leads to policy collapse, regardless of the optimizer's sophistication.

This chapter details the architecture of our reward system, focusing on the methodology used to construct and orchestrate feedback. We move beyond static evaluation to describe a dynamic ecosystem that integrates conflicting objectives—visual fidelity, temporal coherence, and semantic alignment—into robust learning signals~\citep{yu2020gradient}.

\subsection{Heterogeneous Signal Integration}
\label{sec:signal-integration}

Video generation demands the simultaneous satisfaction of orthogonal constraints that cannot be captured by a single monolithic model. Our system treats reward modeling as a signal integration challenge, synthesizing feedback from three distinct representational modalities. This multi-modal approach ensures that optimization is driven by specialized, high-fidelity critics rather than bottlenecked by a single proxy.

\paragraph{1. Semantic Alignment Signals (VLM-Driven).}
To capture high-level instruction following, we deploy large-scale Video-Language Models (VLMs) as semantic judges~\citep{lin2024video,li2025videochat}. Unlike standard CLIP~\citep{radford2021learning} scores which operate in a joint embedding space, these VLMs perform fine-grained reasoning—verifying object counts, spatial relations, and action sequences. The engineering challenge lies in extracting calibrated probabilities from these large models~\citep{he2024videoscore} to provide sufficiently dense gradients for the generator.

\paragraph{2. Temporal Dynamics and Physics Signals.}
Unique to the video domain, we integrate signals derived from dedicated motion estimators~\citep{wu2023tune,singer2022make}. These critics do not judge ``quality" in the abstract, but enforce physical plausibility—penalizing non-causal transitions, static frame repetition, and unnatural warping~\citep{blattmann2023align}. This serves as a critical regularizer against the ``slide-show" failure mode common in diffusion models~\citep{videodiffusionmodels}, balancing the pursuit of semantic alignment with the constraints of physical realism.

\paragraph{3. Perceptual and Safety Signals.}
At the foundational level, we maintain a battery of low-latency critics for artifact detection (blur, noise, compression)~\citep{mittal2012no,zhang2018unreasonable} and safety compliance~\citep{schramowski2023safe}. These operate as ``gating" signals: they impose strict boundaries on the feasible policy space, preventing the model from degenerating into adversarial noise patterns that might otherwise maximize high-level VLM scores.

\section[Part IV: Infrastructure and Training Operations]{Part IV: Infrastructure and Training Operations\protect \footnote{codes are available at https://github.com/Tele-AI}}
\label{sec:infra}

In video generation post-training, preference-based training frameworks face significant infrastructure bottlenecks—manifesting differently in DPO and GRPO. In DPO, the joint backward pass over chosen and rejected trajectories causes high peak memory usage. In GRPO, two inefficiencies arise: first, the VLM-based reward model is deployed on GPUs disaggregated from those used for rollouts and actor updates, yet the strict sequential execution of rollout, reward, and actor leaves GPUs idle during stage transitions; second, in joint reward settings, the Ray-based \cite{ray_paper} framework evaluates multiple lightweight reward workers serially, failing to saturate GPU compute capacity and unnecessarily prolonging reward evaluation.

To address these issues, we design memory--efficient backpropagation, co-located reward execution, and parallelized reward scheduling--enabling scalable, high-throughput preference optimization for video generation.

\subsection{A Ray-based Resource-Efficient Parallel Framework for GRPO}\label{subsec:grpo_mps_reward}

% GRPO supports a reward stage composed of multiple \textit{Reward Workers} (e.g., Aesthetic Predictor, RAFT, VideoCLIP), whose outputs are aggregated to form the final reward for each sampled trajectory.
% In practice, the reward stage becomes a system bottleneck when the worker stack grows, motivating a parallel execution strategy that improves GPU utilization and shortens the reward makespan.

% GRPO supports a reward stage composed of multiple \textit{Reward Workers} (e.g., Aesthetic Predictor, RAFT, VideoCLIP), whose individual outputs are aggregated to form the final reward for each sampled trajectory.
% In practice, the standard serial implementation of this reward stage is computationally inefficient. To address this bottleneck, we propose a parallel execution strategy that accelerates multi-reward computation.

% \subsubsection{GRPO Reward Stage with Multiple Workers}
% Let the composite reward be
% \begin{equation}
% R(x,y) = \sum_{i=1}^{N} \lambda_i \cdot r_i(x,y),
% \end{equation}
% where each $r_i$ is an independent reward worker that consumes the same input $(x,y)$ and produces a scalar score.
% Since these workers are logically parallel, the reward stage performance is primarily determined by runtime orchestration rather than algorithmic dependency.

The heterogeneous reward system of GRPO described in Section~\ref{sec:reward-modeling} poses unique challenges to the infrastructure, and the current implementation is computationally inefficient for two primary reasons.

First, the VLM-based reward model is deployed as an isolated service with GPUs that are disaggregated from those used for rollouts and actor updates. Because rollout generation, reward computation, and actor updates form a strictly sequential dependency chain, this design inevitably leads to GPU idle time—as illustrated in the upper-left panel of Figure~\ref{fig:mps_grpo}. This issue is particularly pronounced in video post-training settings, where video generation is computationally expensive and typically involves long-horizon generation.

Second, in the joint reward scenario, reward computation is implemented as a predefined workflow that invokes all reward workers \emph{serially}. However, as these reward models are typically lightweight, and a single worker cannot fully saturate the available GPU computational resources, as illustrated in the upper-right panel of Figure~\ref{fig:mps_grpo}. Consequently, this leads to prolonged latency in the multi-reward stage.

To address these inefficiencies, we propose a dual optimization strategy: temporal multiplexing consolidates GPU resources across training stages to eliminate idle time, while NVIDIA MPS enables intra-GPU concurrency to saturate compute during lightweight reward evaluation.

% \subsubsection{The Root Cause: Serial Workflow and Low GPU Occupancy}
% In the standard Ray-based \textit{Verl} framework, reward computation is implemented as a predefined Workflow that invokes all workers \emph{serially}. 
% The training runtime enforces a strict workflow chain, making the reward-stage latency approximately additive.
% This is particularly inefficient because reward models are typically lightweight (e.g., VideoCLIP $0.42$B, Aesthetic Predictor $0.73$B, RAFT $5.3$M); 
% These workers impose low GPU load, yet their execution is largely \textbf{compute-bound}, making them good candidates for parallel co-execution.

% \subsubsection{Opportunities for parallel acceleration}
\subsubsection{Temporal Multiplexing and Node Consolidation}

To address the inefficiencies inherent in traditional disaggregated reward systems, we move away from architectures that maintain independent and isolated node pools for Reward (VLM/Joint) components. Instead, we adopt a unified temporal multiplexing approach that enables more effective utilization of computational resources across the entire training lifecycle.

\begin{figure}[t]
    \centering
    \includegraphics[width=0.95\linewidth]{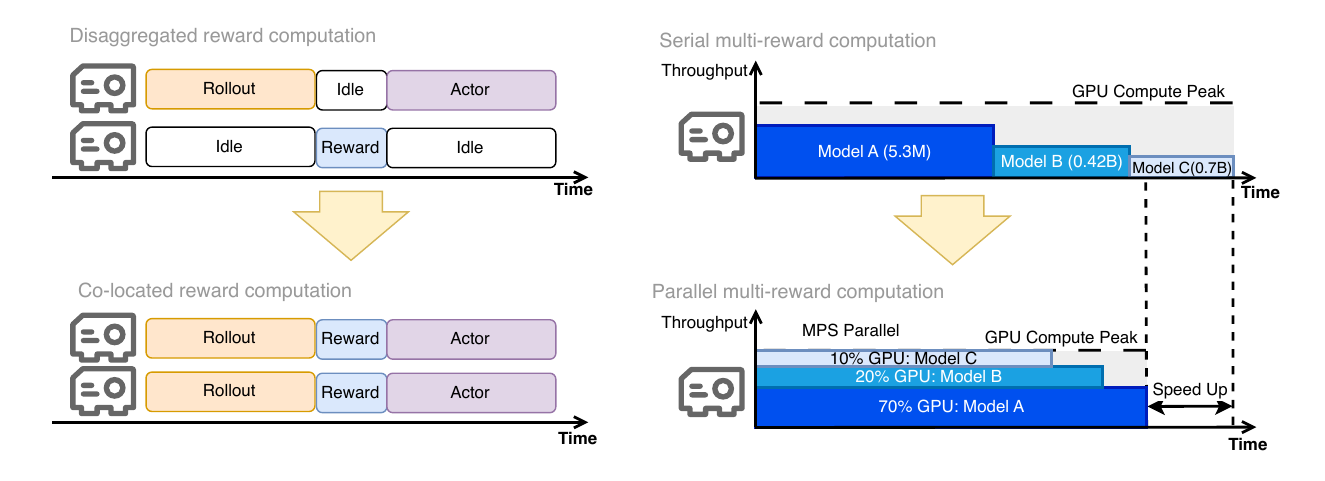}
    \caption{A Ray-Based Resource-Efficient GRPO Framework: Transforming Disaggregated and Serial Reward Computation into Co-Located, Parallel Execution.}
    \label{fig:mps_grpo}
\end{figure}

Rather than statically assigning hardware to fixed functional roles, we leverage Ray to orchestrate a monolithic physical cluster in which a single pool of GPU nodes dynamically transitions across different stages of training, as shown in the lower-left panel of Figure \ref{fig:mps_grpo}. The same hardware is sequentially repurposed for rollout, reward inference, and optimization, ensuring that computational resources remain actively engaged instead of sitting idle while waiting on remote process dependencies.

Concretely, once the rollout phase completes, the corresponding Ray Actors immediately trigger VLM inference and joint reward evaluation on the same hardware. By colocating these stages within a shared resource pool, the system minimizes unnecessary data movement and coordination overhead, thereby improving execution efficiency and overall end-to-end training throughput.

\subsubsection{Accelerated Joint Reward via NVIDIA MPS}
% We eliminate this inefficiency by enabling concurrent execution of heterogeneous reward workers on the same GPU.
% Concretely, we use \textit{threads} to dispatch worker execution commands simultaneously, allowing independent workers to overlap in time rather than waiting in a strict chain.

% To make colocation predictable and efficient, we further leverage NVIDIA MPS to enforce fine-grained compute partitioning across workers: each worker is assigned a tailored GPU compute share according to its workload intensity, preventing interference-driven tail effects while improving overall overlap efficiency.

% Finally, we modify \textit{RayWorkerGroup} to explicitly support overlapping GPU assignments across worker groups, removing the default GPU exclusivity constraint so that multiple workers can legally share a single device under MPS control.

We further optimized the execution of the Joint Reward phase by leveraging NVIDIA Multi-Process Service (MPS)  \cite{mps_paper}. By enabling logical GPU partitioning, MPS allows multiple evaluation operators within the Joint Reward module (such as rule-based scoring and small-model inference) to run concurrently on the same GPU. This fine-grained parallelism significantly boosts throughput and reduces kernel queuing latency during the reward calculation stage--as illustrated in the lower-right panel of Figure\ref{fig:mps_grpo}.

We minimize the reward-stage makespan under grouped parallel execution.
Let $W$ be the reward-worker set, and let
$P=\{p_1,\dots,p_G\}$ be a partition of $W$.
Workers in the same group run concurrently, while groups execute sequentially. Let $C_w(q_w)$ denote worker $w$'s execution time under assigned quota $q_w$.
For each group, the execution time is determined by the slowest worker:
\begin{equation}
T_g=\max_{w\in p_g} C_w(q_w),\qquad g=1,\dots,G.
\label{eq:group_time}
\end{equation}
Accordingly, the reward-stage makespan equals the sum of execution times across all groups:
\begin{equation}
T_{\text{reward}}=\sum_{g=1}^{G}T_g
=\sum_{g=1}^{G}\max_{w\in p_g} C_w(q_w).
\label{eq:reward_makespan_group}
\end{equation}

To reduce $T_{\text{reward}}$, we dispatch workers concurrently via \textit{threads} and use NVIDIA MPS for compute partitioning.
Each worker is assigned quota $q_w\in(0,1]$ through
\textit{set\_default\_active\_thread\_percentage}.
For each concurrently executed group, quota allocation must satisfy:
\begin{equation}
\sum_{w\in p_g} q_w \le 1,\qquad \forall g=1,\dots,G.
\label{eq:reward_quota_budget}
\end{equation}

We choose $(P,\{q_w\})$ with a lightweight greedy search guided by per-worker scaling profiles (measured runtime under different quotas).
This search jointly determines worker grouping and quota split to minimize Eq.~\eqref{eq:reward_makespan_group} under Eq.~\eqref{eq:reward_quota_budget}.

Finally, we modify \textit{RayWorkerGroup} to allow overlapping GPU assignment, removing default GPU exclusivity so multiple reward workers can legally share one GPU under MPS control. To summarize, this approach offers two major benefits:
\begin{itemize}
    \item Reward Makespan Reduction: transforming the reward stage from serial execution toward overlap-dominated execution substantially reduces GRPO iteration latency when multiple workers are enabled.
    \item Higher GPU Utilization: co-running lightweight reward models converts per-worker compute slack into effective throughput, improving GPU occupancy without requiring additional hardware.
\end{itemize}

\subsection{Memory-Efficient DPO}
The standard implementation of DPO is memory-inefficient, exhibiting higher-than-expected peak memory consumption during backpropagation. This overhead becomes a bottleneck for training large-scale models with extended context lengths, ultimately limiting model performance. To facilitate memory-efficient DPO, we first identify the root cause of the currently implementation and then propose a \textbf{Decoupled Gradient Backpropagation (DGB)} strategy.

% To facilitate the training of large-scale models with extended context lengths, we implement a series of memory-centric optimizations. A primary bottleneck in preference learning is the peak memory consumption during the backpropagation of the Direct Preference Optimization (DPO) objective. To address this, we propose a \textbf{Decoupled Gradient Backpropagation (DGB)} strategy.

\subsubsection{The root cause: Shared Parameter Dependency}
\begin{figure}[t]
    \centering
    \includegraphics[width=0.9\linewidth]{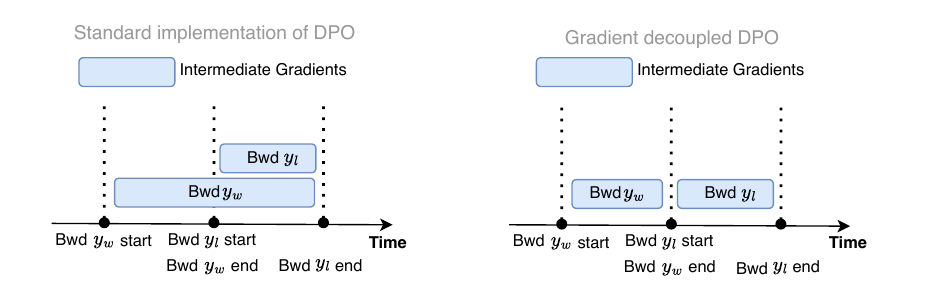}
    \caption{Comparison of Intermediate Gradient Lifespan: Standard Implementation of DPO v.s. Gradient decoupled DPO}
    \label{fig:DPO_decouple}
\end{figure}
% In a conventional DPO implementation, 
% the loss function $\mathcal{L}_{\text{DPO}}$ aggregates the chosen ($y_w$) and rejected ($y_l$) sequences into a single computational graph:

% \begin{align}
% \mathcal{L}_{\text{DPO}}(\theta) = -\mathbb{E}_{(x, y_w, y_l)} \left[ \log \sigma \left( \beta \log \frac{\pi_{\theta}(y_w|x)}{\pi_{\text{ref}}(y_w|x)} - \beta \log \frac{\pi_{\theta}(y_l|x)}{\pi_{\text{ref}}(y_l|x)} \right) \right]\\
% \nabla_{\theta} \mathcal{L}_{\text{DPO}} = \underbrace{- \sigma(-\beta \hat{A}) \cdot \beta}_{\eta} \cdot \left( \nabla_{\theta} \log \pi_{\theta}(y_w|x) - \nabla_{\theta} \log \pi_{\theta}(y_l|x) \right)
% \end{align}

% Although the backward pass processes operators sequentially, a standard unified loss constructs a single \texttt{GraphTask} where model parameters $W \in \theta$ are shared between the chosen and rejected branches. In this configuration, the Autograd engine cannot release the intermediate gradient tensors (e.g., outputs from \texttt{mm\_mat2\_backward}) of the first branch until the second branch also completes its traversal to the same parameter node $W$. This is because the engine must maintain these partial results for gradient accumulation. 

% Consequently, as the backpropagation progresses through the second branch, the new intermediate gradients and activations are allocated while the previous branch's tensors are still resident in memory. This \textbf{temporal overlap of memory residency} creates a peak memory requirement proportional to $O(L_w + L_l)$.
The loss function and gradient of DPO are denoted as follows:
\begin{align}
\mathcal{L}_{\text{DPO}}(\theta) = -\mathbb{E}_{(x, y_w, y_l)} \left[ \log \sigma \left( \beta \log \frac{\pi_{\theta}(y_w|x)}{\pi_{\text{ref}}(y_w|x)} - \beta \log \frac{\pi_{\theta}(y_l|x)}{\pi_{\text{ref}}(y_l|x)} \right) \right]\label{eq.DPO_loss}\\
\mathcal{J}(\theta)=\nabla_{\theta} \mathcal{L}_{\text{DPO}} = - \sigma(-\beta \hat{A}) \cdot \beta \cdot \left( \nabla_{\theta} \log \pi_{\theta}(y_w|x) - \nabla_{\theta} \log \pi_{\theta}(y_l|x) \right)\label{eq.DPO_grad}
\end{align}
where $\hat{A} = \log \frac{\pi_{\theta}(y_w|x)}{\pi_{\text{ref}}(y_w|x)} - \log \frac{\pi_{\theta}(y_l|x)}{\pi_{\text{ref}}(y_l|x)}$ denotes the preference margin. 

The DPO loss function \eqref{eq.DPO_loss} combines two computational branches: one corresponding to the chosen sequence $y_w$ and the other to the rejected sequence $y_l$. In a standard DPO implementation, a single \texttt{GraphTask} is launched during the backward pass, and gradients \eqref{eq.DPO_grad} are computed via a single call to $\mathcal{J}.backward()$. Consequently, the model parameters $W\in \theta$ are shared across both branches. This sharing induces what we refer to as a \textit{shared parameter dependency}. Because the Autograd engine must maintain partial results for gradient accumulation, intermediate gradient tensors from the first branch cannot be released until the backward pass of the second branch completes--as visualized in the left part of Figure \ref{fig:DPO_decouple}. Consequently, peak memory consumption scales with the combined size of intermediate tensors from both branches, yielding a memory complexity of $O(L_w + L_l)$, where $L_w$ and $L_l$ denote the number of such tensors in the chosen and rejected branches, respectively.

\subsubsection{Decoupled Gradient Backpropagation}
% We propose to eliminate this overlap by decoupling the gradient calculation without altering the total compute cost or the mathematical integrity of DPO. Let the preference margin be $\hat{A} = \log \frac{\pi_{\theta}(y_w|x)}{\pi_{\text{ref}}(y_w|x)} - \log \frac{\pi_{\theta}(y_l|x)}{\pi_{\text{ref}}(y_l|x)}$. The total gradient is:

% \begin{equation}
% \nabla_{\theta} \mathcal{L}_{\text{DPO}} = \underbrace{- \sigma(-\beta \hat{A}) \cdot \beta}_{\eta} \cdot \left( \nabla_{\theta} \log \pi_{\theta}(y_w|x) - \nabla_{\theta} \log \pi_{\theta}(y_l|x) \right)
% \end{equation}

% By pre-calculating the scalar weighting coefficient $\eta$ in a \textit{non-gradient} context, we can manually trigger two isolated backward passes:
% \begin{equation}
% \mathcal{J}_{w}(\theta) = -\eta \cdot \sum \log \pi_{\theta}(y_w|x), \quad \mathcal{J}_{l}(\theta) = \eta \cdot \sum \log \pi_{\theta}(y_l|x)
% \end{equation}
Our key insight for alleviating the shared parameter dependency is gradient backpropagation decoupling. This approach reduces peak memory consumption while preserving both the total computational cost and the mathematical correctness of the standard DPO implementation. Specifically, the gradient of the DPO loss \eqref{eq.DPO_grad} can be decomposed as follows:
\begin{align}
\begin{aligned}
\mathcal{J}(\theta)=\nabla_{\theta} \mathcal{L}_{\text{DPO}} =& - \underbrace{\sigma(-\beta \hat{A}) \cdot \beta}_{\eta} \cdot \left( \nabla_{\theta} \log \pi_{\theta}(y_w|x) - \nabla_{\theta} \log \pi_{\theta}(y_l|x) \right)\\
=&-\eta \cdot \nabla_{\theta} \log \pi_{\theta}(y_w|x)+\eta \cdot \nabla_{\theta} \log \pi_{\theta}(y_l|x)\label{eq.DPO_grad_decoupled}
\end{aligned}
\end{align}
where $\eta=\sigma(-\beta \hat{A}) \cdot \beta$ is scalar weighting coefficient.

Crucially, $\eta$ is a scalar that can be pre-computed  in a \textit{non-gradient} context. Leveraging the decoupled form \eqref{eq.DPO_grad_decoupled}, we then perform two separate backward passes: 

\begin{equation}
\mathcal{J}_{w}(\theta) = -\eta \cdot \nabla_{\theta} \log \pi_{\theta}(y_w|x), \quad \mathcal{J}_{l}(\theta) = \eta \cdot \nabla_{\theta} \log \pi_{\theta}(y_l|x)
\end{equation}
This enables us to execute $\mathcal{J}_{w}.backward()$ and $\mathcal{J}_{l}.backward()$ independently--each launching its own \texttt{GraphTask}. As a result, the Autograd engine can immediately release intermediate tensors after each backward pass--as illustrated in the right part of Figure \ref{fig:DPO_decouple}--thereby significantly reducing peak memory usage without altering the final accumulated gradients

To summarize, this approach offers two major advantages:
\begin{itemize}
    \item \textbf{Peak Memory Reduction}: The memory complexity is reduced from $O(L_w + L_l)$ to $O(\max(L_w, L_l))$, effectively preventing Out-of-Memory (OOM) errors without requiring additional GPU resources.
    \item \textbf{Zero Computational Overhead}: Since the total number of floating-point operations (FLOPs) remains identical to the standard DPO backward pass, this optimization achieves significant memory savings without sacrificing \textit{ training throughput}.
\end{itemize}

% \subsubsection{Performance and Memory Benefits}
% By executing $\mathcal{J}_{w}.backward()$ and $\mathcal{J}_{l}.backward()$ as two separate execution invocations, we force the Autograd engine to finalize the \texttt{GraphTask} and release all branch-specific intermediate tensors immediately after each call. 

% This approach offers two major advantages:
% \begin{itemize}
%     \item \textbf{Peak Memory Reduction}: The memory complexity is reduced from $O(L_w + L_l)$ to $O(\max(L_w, L_l))$, effectively preventing Out-of-Memory (OOM) errors without requiring additional GPU resources.
%     \item \textbf{Zero Computational Overhead}: Since the total number of floating-point operations (FLOPs) remains identical to the standard DPO backward pass, this optimization provides significant memory relief with \textit{no loss in training throughput}.
% \end{itemize}

\section{Evaluation}
\label{sec:evaluation}

This chapter reports experimental validation of our post-training framework, with an emphasis on \emph{reproducibility} and \emph{attribution}. We separate evaluation into three complementary parts: (i) subjective human scoring, (ii) qualitative demonstrations, and (iii) objective benchmarks and ablations. The first two parts are intentionally reserved for later insertion, as they depend on ongoing data collection and curated media assets.

\subsection{Subjective Evaluation I: Human Scoring}
\label{sec:human-eval}

Automatic metrics provide scalable and reproducible signals for post-training, but they remain limited in capturing holistic perceptual qualities such as temporal naturalness, semantic coherence, and overall viewing experience.
To complement automatic evaluation, we conduct a controlled human study to assess relative preference between our method and a strong image-to-video baseline, Wan2.2-14B I2V~\citep{wang2025wan}.

Following recent large-scale video generation reports, including LongCat~\citep{team2025longcat} and SeaDance, we adopt a \emph{Good--Same--Bad (GSB)} comparison protocol.
GSB is well-suited for video evaluation, as it explicitly allows annotators to express indifference when differences are subtle, thereby reducing forced or noisy decisions in marginal cases.

\paragraph{Evaluation setup.}
Human evaluation is conducted in a pairwise comparison setting.
For each prompt, videos generated by our method and Wan2.2-14B are presented side-by-side under a fixed rendering envelope, including identical resolution, frame rate, and duration.
All videos are synchronized to start at the same frame, and the left--right order is randomized to avoid positional bias.
Each comparison focuses on a single evaluation aspect to reduce cognitive load and cross-aspect interference.

\paragraph{Evaluation aspects.}
We evaluate five complementary aspects that jointly characterize video generation quality:
\begin{itemize}
    \item \textbf{Visual quality}: overall appearance, sharpness, and absence of visual artifacts;
    \item \textbf{Motion quality}: temporal coherence, smoothness, and plausibility of motion patterns;
    \item \textbf{Text alignment}: consistency between the generated video and the input prompt semantics;
    \item \textbf{Preservation}: stability of subjects, identities, and key content across time;
    \item \textbf{Overall}: holistic preference considering all factors jointly.
\end{itemize}

\paragraph{GSB metrics.}
Table~\ref{tab:gsb_human_eval} reports three complementary statistics derived from GSB annotations.
\emph{WinRate} is computed as $G/(G+B)$ and reflects the fraction of decisive wins, ignoring ties.
\emph{Preference} is defined as $(G + 0.5S)/(G+S+B)$, which accounts for neutral judgments by assigning half credit to ``Same''.
\emph{Margin}, defined as $(G-B)/(G+S+B)$, captures the signed preference gap and provides a conservative estimate of preference strength.
Together, these metrics provide a robust and interpretable summary of human judgments.

\paragraph{Results and analysis.}
As shown in Table~\ref{tab:gsb_human_eval}, our method consistently outperforms Wan2.2-14B across all five aspects.
The improvements are most pronounced for motion quality and text alignment, where preference margins exceed $24\%$.
This indicates that annotators not only prefer our results more frequently, but also do so with a clear and confident margin, reflecting substantial gains in temporal coherence and semantic correctness.

Visual quality and preservation also show stable improvements, with win rates above $58\%$ and positive margins.
These results suggest that the improvements in motion and alignment do not introduce noticeable regressions in appearance quality or identity consistency.
In particular, the preservation scores indicate that our method maintains subject stability over time, rather than trading off consistency for more expressive motion.

For the overall criterion, our method achieves a $71.18\%$ win rate and a $32.71\%$ margin against a strong large-scale I2V baseline.
This strong overall preference reflects cumulative improvements across multiple perceptual dimensions, rather than dominance by a single factor.

\begin{table}[!htp]
\centering
\small
\begin{tabular}{lccc}
\toprule
\textbf{Evaluation Aspect} & \textbf{WinRate} $\uparrow$ & \textbf{Preference} $\uparrow$ & \textbf{Margin} $\uparrow$ \\
\midrule
Visual quality   & 58.71\% & 52.17\% & 4.35\% \\
Motion quality   & 70.72\% & 62.47\% & 24.90\% \\
Text alignment   & 77.39\% & 62.15\% & 24.13\% \\
Preservation     & 63.28\% & 54.15\% & 8.15\% \\
Overall          & 71.18\% & 66.38\% & 32.71\% \\
\bottomrule
\end{tabular}
\caption{
GSB human evaluation results comparing our method against \textbf{Wan2.2-14B I2V}.
WinRate is computed as $G/(G+B)$, Preference as $(G+0.5S)/(G+S+B)$, and Margin as $(G-B)/(G+S+B)$.
}
\label{tab:gsb_human_eval}
\end{table}

\paragraph{Discussion.}
The human evaluation results demonstrate that the proposed method yields perceptible improvements over a strong large-scale image-to-video baseline, particularly in motion coherence and prompt alignment.
The consistency of gains across all evaluated aspects suggests that the improvements are balanced and systematic, rather than driven by isolated visual effects.

\begin{figure}[!htp]
	\centering
	\includegraphics[width=0.45\linewidth]{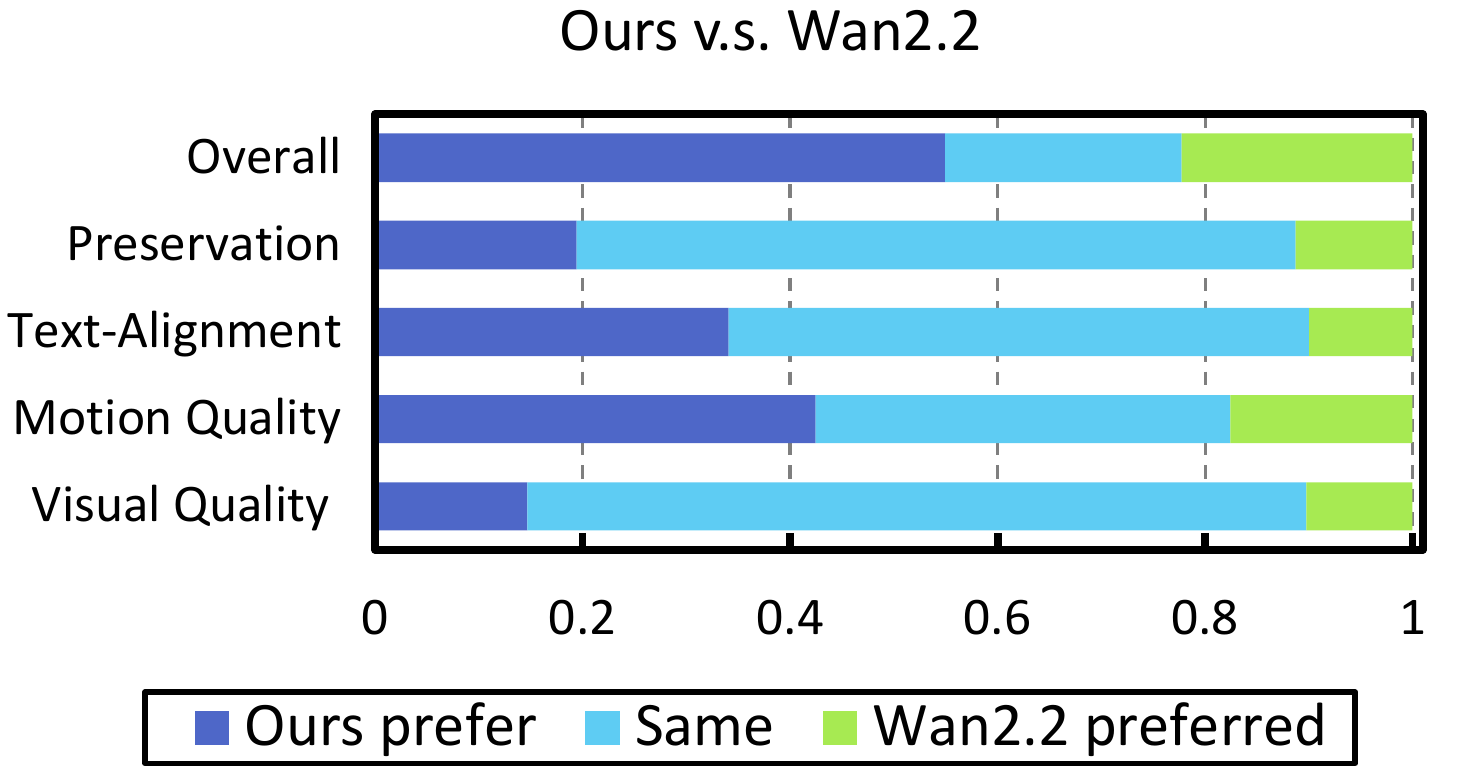}
	\caption{
	Visualization of GSB human evaluation results comparing our method with {Wan2.2-14B I2V}.
	Bars indicate the proportions of Good, Same, and Bad judgments for each evaluation aspect.
	The dominance of ``Good'' responses in motion quality, text alignment, and overall preference highlights the consistency of human judgments.
	}
	\label{fig:gsb}
\end{figure}

\subsection{Subjective Evaluation II: Demo Gallery}
\label{sec:qualitative-demos}

To complement controlled human evaluation and quantitative analysis, we present a curated demo gallery illustrating representative generation capabilities of our post-training framework built upon Wan2.2. The purpose of this section is to provide qualitative evidence of model behavior in realistic, user-facing scenarios that are difficult to fully capture with scalar metrics.
All examples shown are generated by the same post-trained model under consistent decoding settings.

\paragraph{Gallery A: Character fidelity and motion naturalness.}
Character-centric prompts constitute a high-frequency and high-sensitivity usage scenario for video generation systems. Common challenges include identity drift, anatomical inconsistency, and temporal jitter, particularly under long-horizon generation or camera motion.

\begin{figure*}[!htp]
    \centering
    \includegraphics[width=\linewidth]{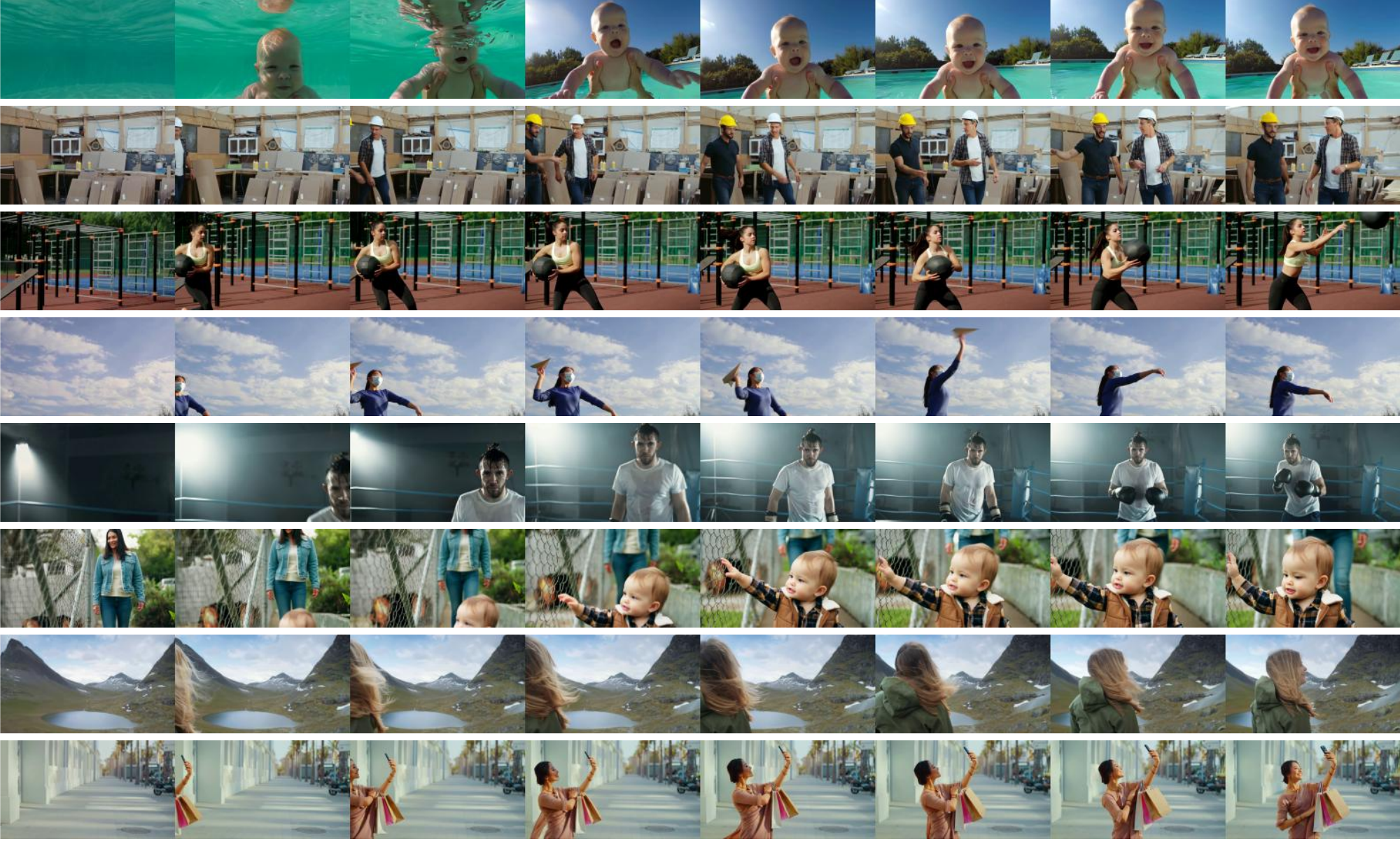}
    \caption{
    \textbf{Demo Gallery A: Character fidelity and motion naturalness.}
    Representative character-centric generations produced by our post-trained Wan2.2 model,
    demonstrating stable identity, coherent articulation, and temporally smooth motion across
    diverse actions and viewpoints.
    }
    \label{fig:demo_characters}
\end{figure*}

Figure~\ref{fig:demo_characters} presents representative character-generation results produced by our post-trained model. Across diverse actions and viewpoints, generated characters exhibit stable identity, coherent articulation, and temporally smooth motion. Facial structure and body proportions remain consistent across frames, and motion trajectories appear continuous and physically plausible, even in sequences involving rapid movement or camera dynamics.

\paragraph{Gallery B: Scene realism and sensory visual quality.}
In addition to subject-centric content, practical deployment requires robust generation quality across a wide range of environments, including natural landscapes, urban scenes, and indoor settings. Such scenarios demand consistent global structure, stable lighting, and high perceptual fidelity over time.

\begin{figure*}[!htp]
    \centering
    \includegraphics[width=\linewidth]{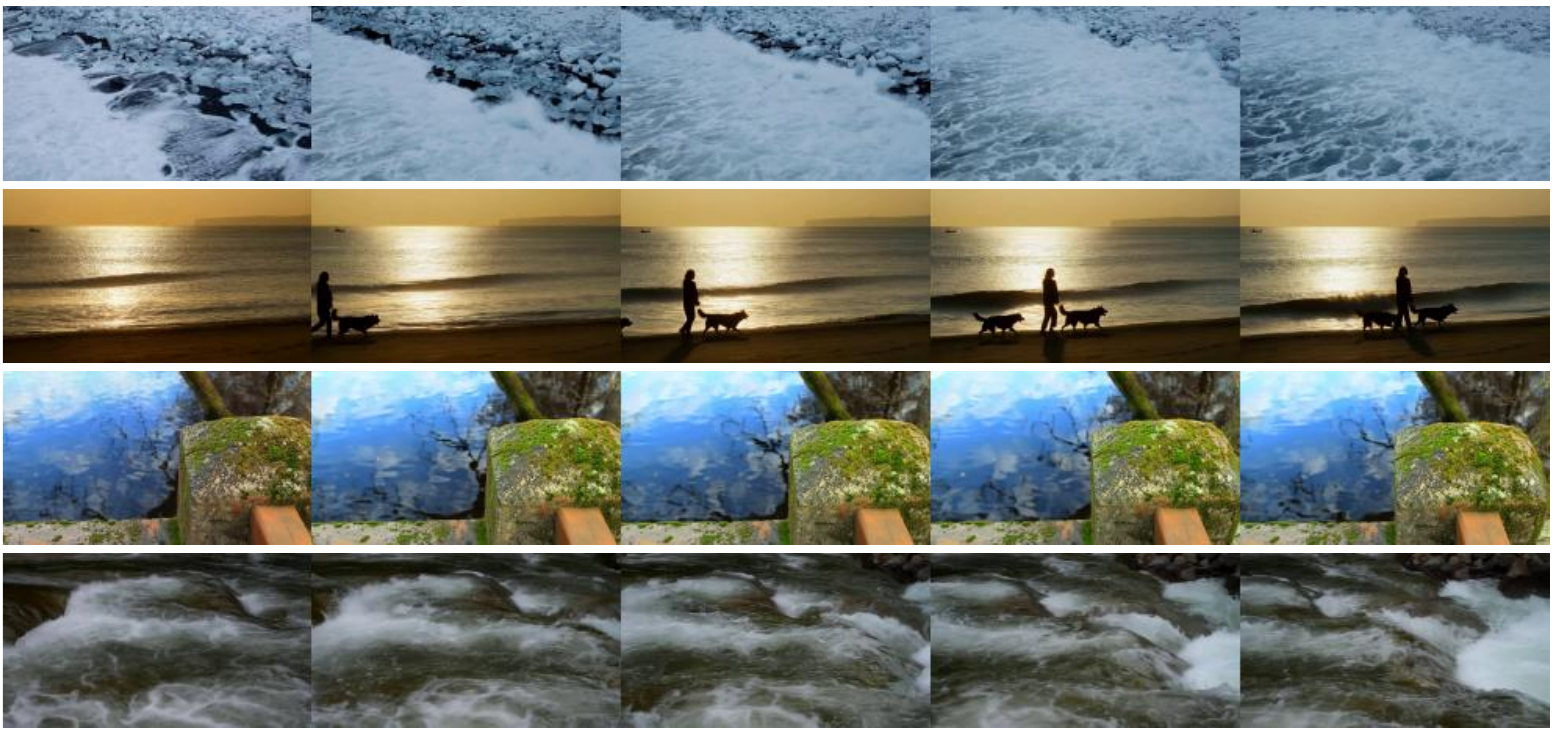}
    \caption{
    \textbf{Demo Gallery B: Scene realism and sensory visual quality.}
    Representative scene-generation results from the post-trained Wan2.2 model, illustrating
    high visual fidelity, stable lighting, and coherent global structure across a range of natural
    and indoor environments.
    }
    \label{fig:demo_scenery}
\end{figure*}

Figure~\ref{fig:demo_scenery} showcases representative scenery-focused generations from our post-trained model. The results demonstrate strong photometric consistency, clear texture detail, and coherent global composition across frames. Temporal stability is maintained without noticeable flicker or abrupt appearance shifts, supporting visually convincing and cinematic scene generation under diverse conditions.

\paragraph{Gallery C: Physics-related phenomena.}
Natural phenomena such as smoke, water, fog, and specular or reflective surfaces pose particular challenges for video generation models due to their complex, view-dependent dynamics and strong temporal coupling. Errors in these cases often manifest as non-physical motion, inconsistent directionality, unstable transparency, or implausible reflection behavior, which are visually salient and difficult to correct post hoc.

\begin{figure*}[!htp]
    \centering
    \includegraphics[width=\linewidth]{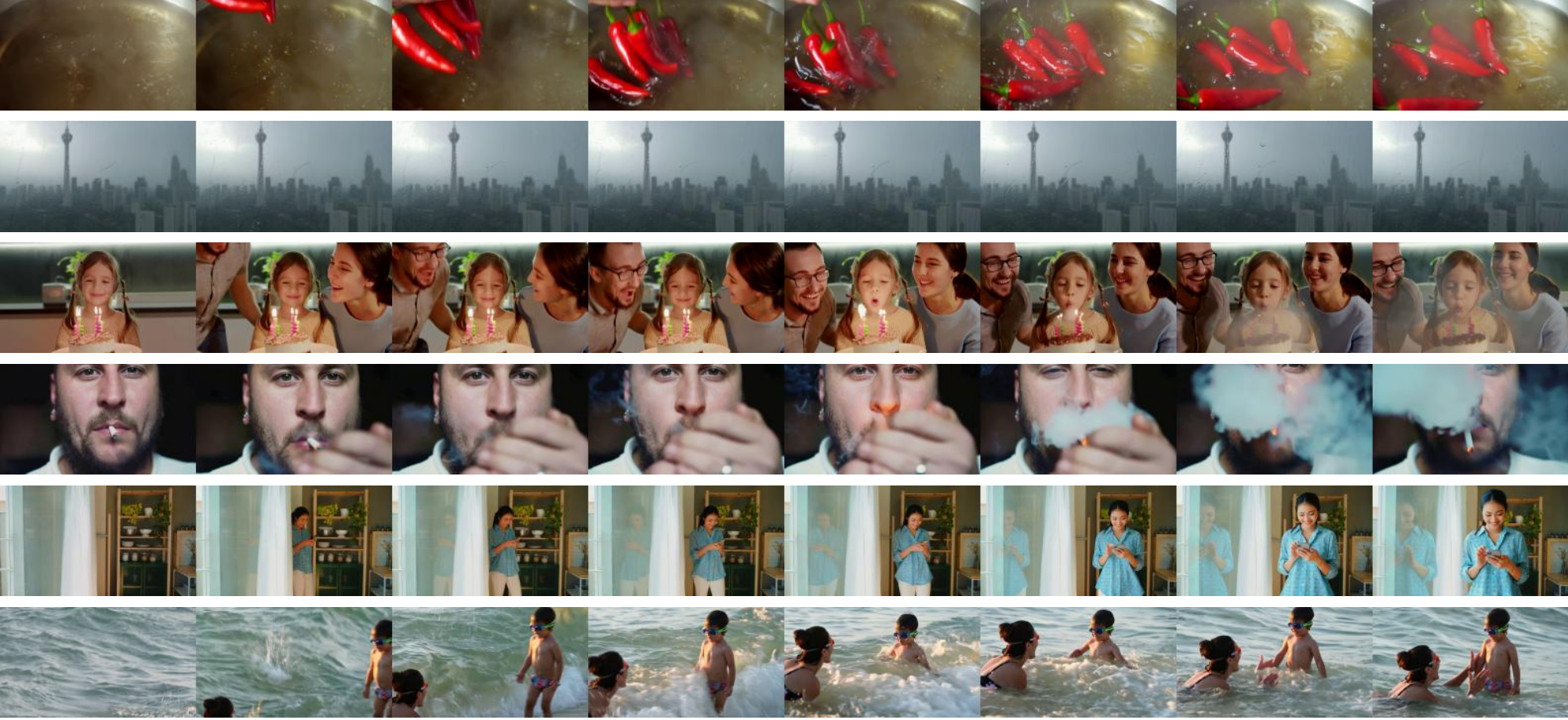}
    \caption{
    \textbf{Demo Gallery C: Physics-related phenomena (smoke, water, fog, and specular reflection).}
    Representative generations from the post-trained Wan2.2 model, demonstrating temporally
    coherent and visually plausible dynamics for smoke, water, and fog effects.
    }
    \label{fig:demo_physics}
\end{figure*}

Figure~\ref{fig:demo_physics} presents representative generations involving smoke, water, fog, and specular reflection effects (e.g., glass- or mirror-like surfaces). The model produces smooth and temporally coherent evolution patterns, with stable directionality for fluid-like motion and consistent reflection behavior under viewpoint changes. We observe reduced temporal discontinuities, fewer hallucinated structures, and improved stability in view-dependent appearance, resulting in visually plausible representations across a range of physics-driven and optically-sensitive scenarios.

\paragraph{Summary.}
Across character-centric content, diverse scenes, and physics-related phenomena, the demo gallery highlights the effectiveness of our post-training framework in producing stable, coherent, and high-quality video generation. These qualitative results are consistent with the design goals of the staged post-training pipeline: establishing structural stability during supervised fine-tuning, refining measurable properties via reinforcement learning, and consolidating overall perceptual quality through preference-based optimization.

\subsection{Objective Evaluation: Benchmarks and Ablations}
\label{sec:objective-eval}

Objective evaluation is designed to be consistent with the system decomposition described earlier. We report both \textbf{in-domain} metrics (aligned with the reward interfaces used during training) and \textbf{out-of-domain} benchmarks (to quantify generalization).

\subsubsection{Spatial-Structure-Aware SFT: Spatial and Geometric Stability.}
\begin{figure*}[!htp]
    \centering
    \includegraphics[width=0.75\linewidth]{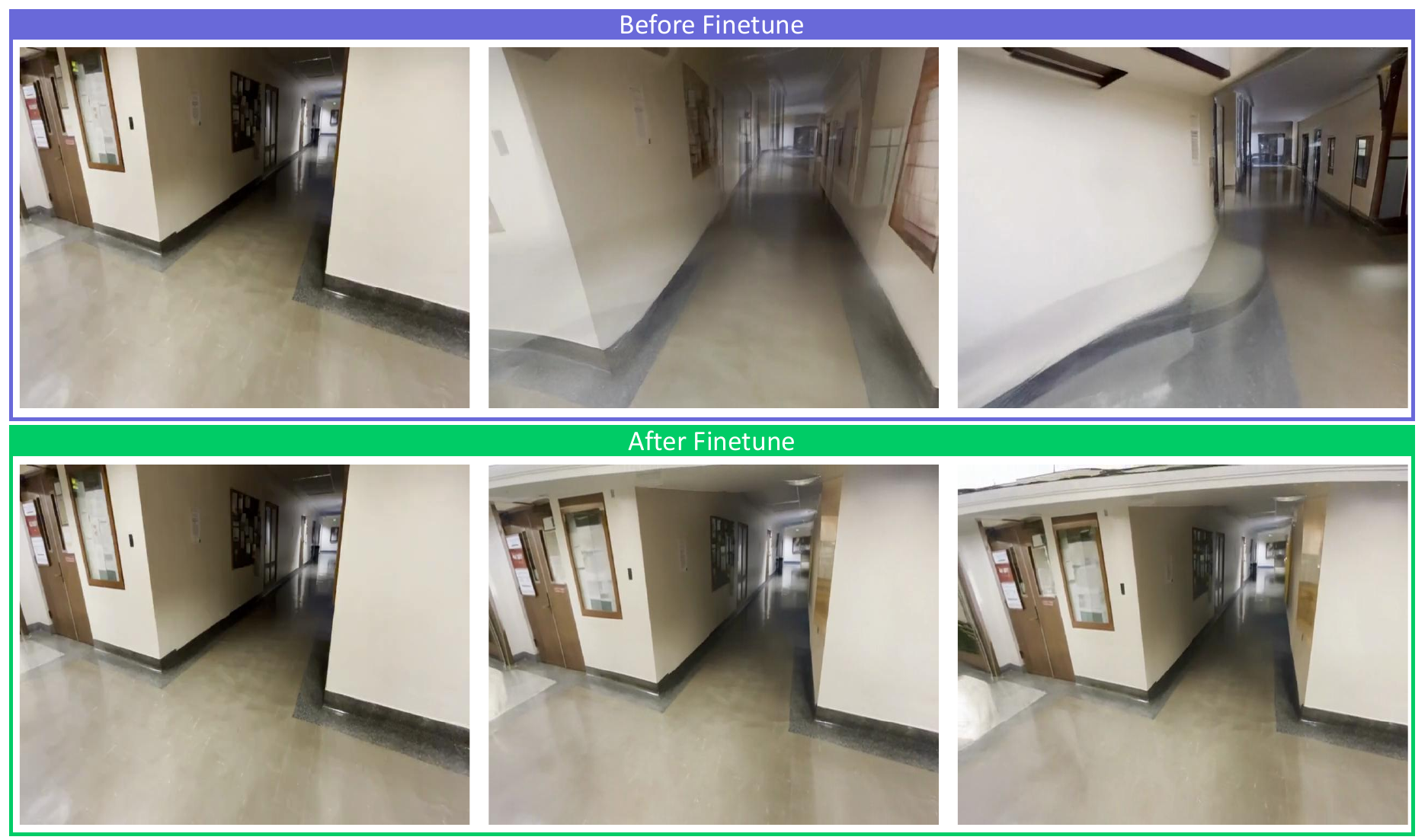}
    \caption{
    Global spatial stability under lateral camera motion.
    The top row shows results from the base model, while the bottom row corresponds to the
    reward-weighted SFT model.
    The baseline exhibits spatial collapse and background misalignment as camera motion progresses,
    whereas the reward-weighted model preserves coherent global geometry and consistent depth
    ordering.
    }
    \label{fig:vggt_sft_demo1}
\end{figure*}
\begin{figure*}[!htp]
    \centering
    \includegraphics[width=0.745\linewidth]{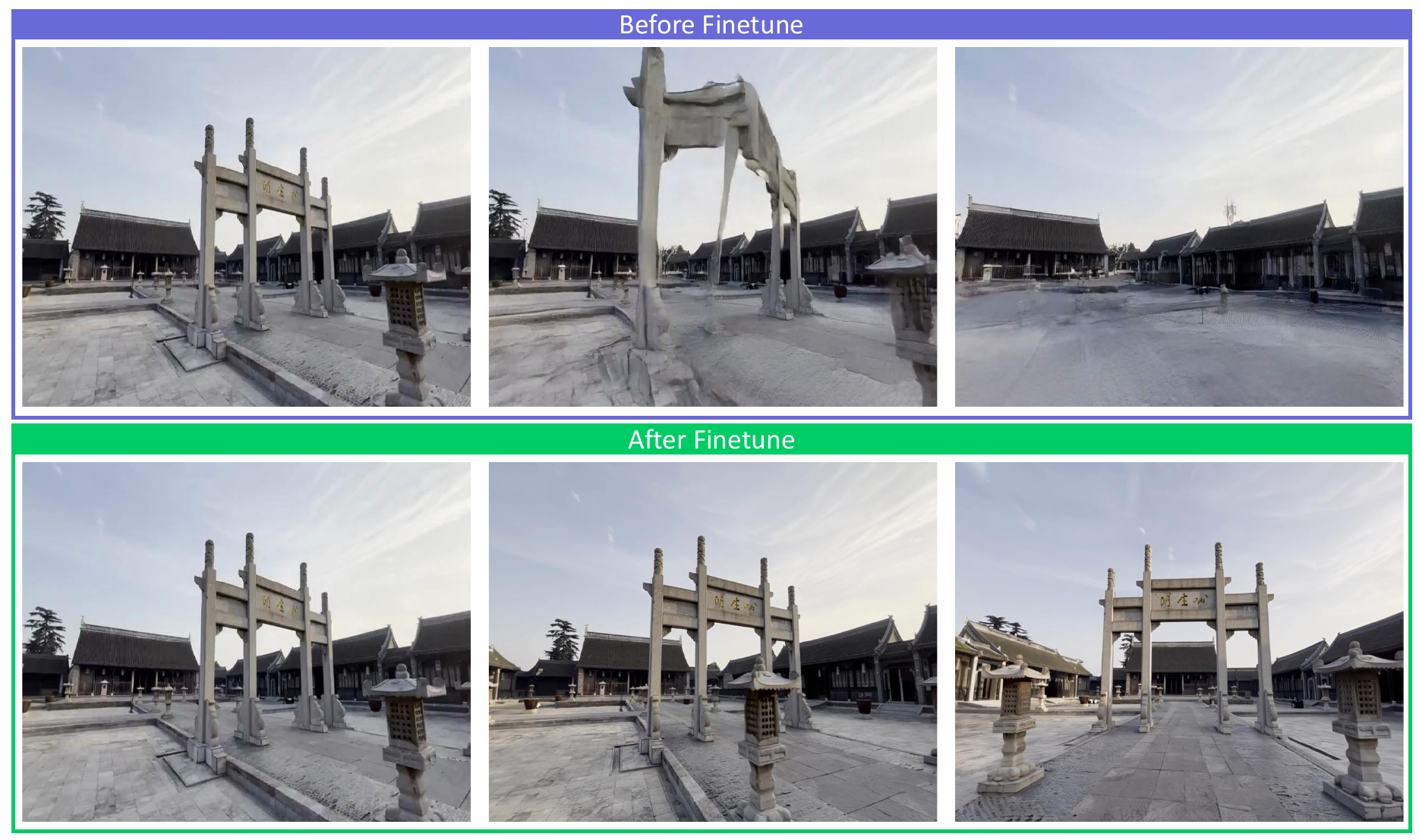}
    \caption{
    Object-level geometric consistency under close-range and orbiting views.
    The base model (top row) suffers from shape distortion and object instability under large
    viewpoint changes.
    The reward-weighted SFT model (bottom row) maintains rigid object structure and clean,
    temporally stable boundaries.
    }
    \label{fig:vggt_sft_demo2}
\end{figure*}

\paragraph{Global spatial stability under camera motion.}
Figure~\ref{fig:vggt_sft_demo1} compares the base model and the reward-weighted SFT model under lateral camera motion. The base model exhibits progressive spatial collapse and misalignment in background structures, often accompanied by non-physical relative drift between foreground and background regions as the viewpoint changes. These errors accumulate over time and lead to noticeable degradation of global scene geometry.

In contrast, the reward-weighted SFT model maintains a coherent global layout throughout the sequence. Background structures remain spatially stable, depth ordering is preserved, and relative motion between scene elements is consistent with the camera trajectory. This demonstrates that structure-aware supervision during SFT effectively stabilizes global geometry under challenging viewpoint changes.

\paragraph{Object-level geometric consistency under close-range views.}
Figure~\ref{fig:vggt_sft_demo2} focuses on object-level behavior under close-range and orbiting camera motion. In the baseline results, such viewpoints frequently induce shape distortion, partial object disappearance, and localized visual artifacts, reflecting weak geometric constraints during generation.

The reward-weighted SFT model substantially mitigates these failure modes. Object shapes remain rigid and well-defined across frames, boundaries are clean and temporally stable, and geometric consistency is preserved even under large viewpoint changes. These results indicate that the proposed SFT strategy improves fine-grained geometric robustness, complementing its effect on global spatial stability.

\subsubsection{Physics-Aware SFT: Fluid Dynamics Slice}
\label{sec:eval-physics}
\begin{figure*}[!htp]
	\centering
	\includegraphics[width=\linewidth]{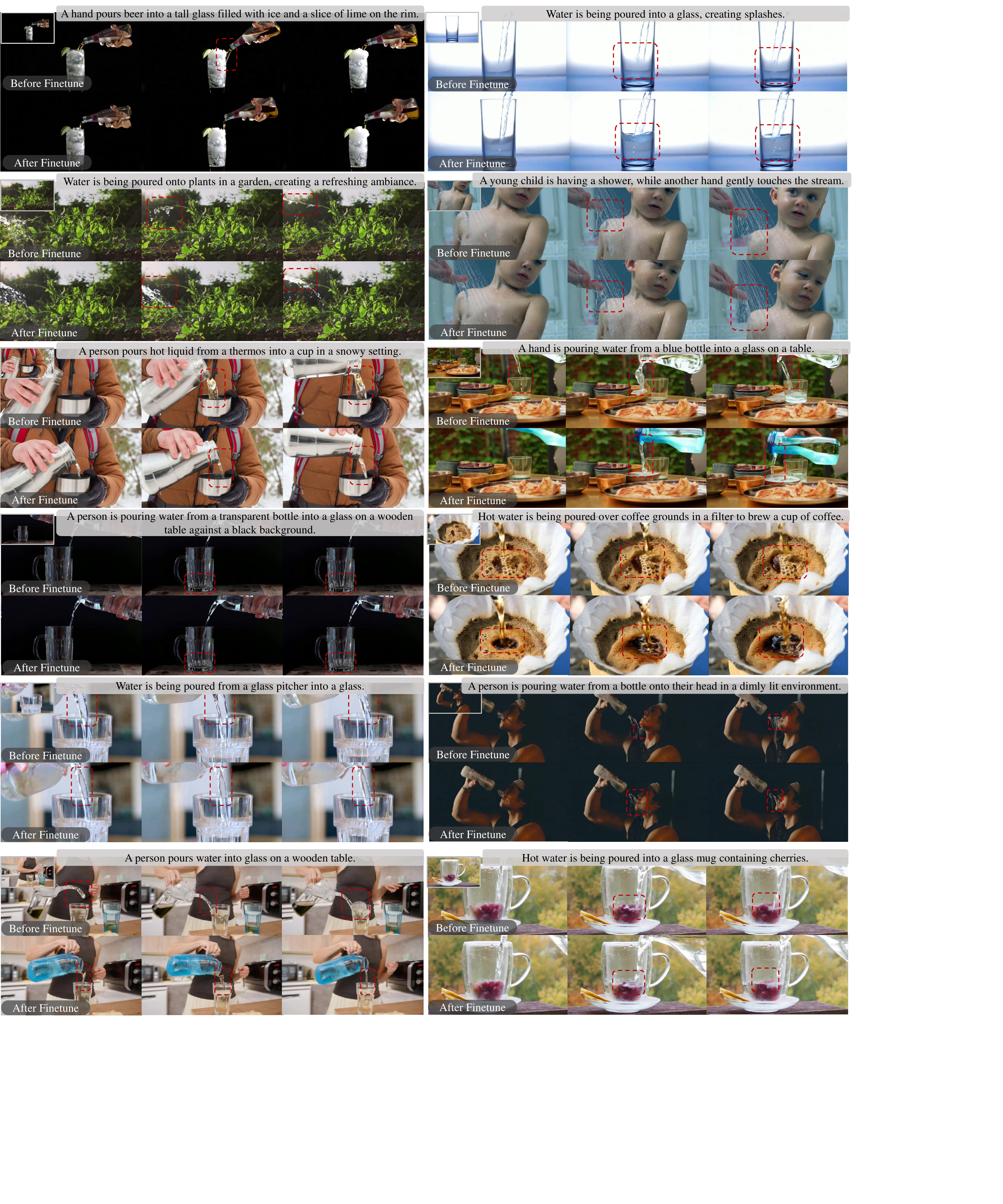}
	\caption{
	Qualitative comparison of image-to-video generation on fluid-related prompts.
	Compared to the baseline Wan model and other video generation methods, our physics-aware SFT produces more temporally coherent fluid motion, with smoother flow trajectories and fewer physically implausible artifacts.
	}
	\label{fig:physic_simu_result}
\end{figure*}

This subsection evaluates the physics-aware SFT slice introduced in Stage~I for \textbf{image-to-video (I2V)} generation.
The objective is to assess whether auxiliary optical-flow supervision improves the temporal plausibility and stability of fluid motion when generating videos conditioned on initial visual states.

\paragraph{Evaluation setting.}
We evaluate motion consistency on two complementary test sets:
(i) \textbf{real-world fluid videos}, consisting of natural fluid phenomena (e.g., pouring, spilling) collected from stock footage;
and (ii) \textbf{simulation-generated fluid videos}, which exhibit controlled but physically grounded fluid dynamics.
For both settings, optical flow is extracted using RAFT and treated as a pseudo ground truth for quantitative comparison.
This evaluation focuses on motion behavior rather than appearance fidelity, and is therefore specific to I2V generation.

\begin{table}[!htp]
	\centering
	\small
	\caption{Optical flow prediction metrics on fluid-effect video test sets. Lower values indicate more coherent and stable motion.}
	\label{tab:physics-flow}
	\begin{tabular}{c|cccc}
		\toprule
		\textbf{Test Set} & \textbf{EPE (all)$\downarrow$} & \textbf{$>$1px$\downarrow$} & \textbf{$>$3px$\downarrow$} & \textbf{F1-all$\downarrow$} \\
		\midrule
		Real-world fluid effects   & 0.538 & 14.7\% & 0.0\%  & 4.680  \\
		Simulation fluid effects   & 1.541 & 21.8\% & 10.0\% & 10.040 \\
		\bottomrule
	\end{tabular}
\end{table}

\paragraph{Metric interpretation.}
End-Point Error (EPE) measures the average discrepancy between predicted and reference optical flow vectors, capturing deviations in motion magnitude and direction.
The $>$1px and $>$3px metrics quantify the proportion of pixels with large motion errors, while F1-all summarizes overall flow inconsistency under a standard threshold.
In the context of video generation, these metrics serve as proxies for temporal smoothness and physical coherence, rather than absolute flow accuracy.

\paragraph{Results and analysis.}
As shown in Table~\ref{tab:physics-flow}, the physics-aware SFT model achieves low flow errors on both real-world and simulation-based fluid videos.
Performance on real-world fluid effects indicates that the model learns motion patterns that generalize beyond synthetic settings, while results on simulation fluid effects demonstrate its ability to capture structured, physically meaningful flow fields.
Across both test sets, lower EPE and error rates correspond to fewer temporal discontinuities and more stable fluid motion over time.

\paragraph{Qualitative comparison.}
Figure~\ref{fig:physic_simu_result} provides qualitative comparisons on representative I2V examples.
Videos generated with physics-aware SFT exhibit smoother liquid surfaces, consistent motion directionality, and reduced flicker or hallucinated splashes, whereas baseline models tend to produce fragmented or temporally inconsistent fluid behavior. 
These results support the role of physics-aware SFT as a targeted enhancement for I2V generation, providing a stronger and more stable initialization for subsequent post-training stages.

\subsubsection{BGPO: Reliability-Aware Optimization}
\label{sec:eval-bgpo}
In this section we evaluate BGPO introduced in Stage II of BPGO on T2V and I2V generation tasks to validate that prior-guided trust allocation enhances training stability under ambiguous reward supervision. Through quantitative comparisons, human evaluation, and ablation studies, we analyze how BGPO's two core components—Reward Adaptation Strategy (RAS) and Confidence-based Reward Transformation (CRT)—contribute to improved alignment and convergence.

\begin{table}[!htp]
    \centering
    \caption{Quantitative comparison on T2V and I2V generation tasks. BPGO consistently outperforms pretrained baselines and GRPO across all metrics, with particularly notable improvements on alignment-sensitive measures. \textdagger Our implemented DanceGRPO.}
    \begin{tabular}{cc|cccc}
    \toprule
        \textbf{Task} & \textbf{Method} & \textbf{VideoClipXL} & \textbf{VideoAlign-TA} & \textbf{VideoAlign-overall} & \textbf{Qwen3-VL-Embedding}\\
        \midrule
        \multirow{3}{*}{T2V}
        & Wan2.1 & 2.6563 & 1.0638 & 0.0939 & 0.6741 \\
        & GRPO\textdagger & 2.6714 & 0.8984 & -0.5411 & 0.6722\\
        & BPGO & 2.6788 & 1.1193 & -0.0478 & \textbf{0.6754} \\
        \midrule
        \multirow{3}{*}{I2V}
        & Wan2.2 & 2.6726 & 1.0633 & -0.7623 & 0.6885\\
        & GRPO\textdagger & 2.0713 & 0.2307 & -1.8932 & 0.4513 \\
        & BPGO & 2.6855 & 1.0589 & -1.0491 & \textbf{0.6890}\\
    \bottomrule
    \end{tabular}
    \label{tab:bgpo-sota}
\end{table}

\paragraph{Evaluation setting.}
In this section, we evaluate BGPO, the core optimization component of Stage II in the BPGO, on T2V and I2V generation tasks. We validate that BPGO's prior-guided trust allocation enhances training stability under ambiguous reward supervision. Through quantitative comparisons, human evaluation, and ablation studies, we analyze how BGPO's two core components—Reward Adaptation Strategy (RAS) and Confidence-based Reward Transformation (CRT)—contribute to improved alignment and convergence in BPGO.

\begin{table}[!htp]
    \centering
    \caption{Fine-grained VBench evaluation on video generation quality. BPGO with RAS+CRT achieves the best performance across 6 out of 9 temporal and spatial consistency metrics.}
    \tabcolsep=0.02cm 
    \begin{tabular}{c|ccccccccc}
    \toprule
        \textbf{Method} & \makecell{\textbf{Object}\\\textbf{Class}} & \makecell{\textbf{Multiple}\\\textbf{Objects}} & \makecell{\textbf{Human}\\\textbf{Action}} & \textbf{Color} & \textbf{Scene} & \makecell{\textbf{Temporal}\\\textbf{Style}} & \makecell{\textbf{Overall}\\\textbf{Consistency}} & \makecell{\textbf{Appearance}\\\textbf{Style}} & \makecell{\textbf{Spatial}\\\textbf{Relationship}}\\
        \midrule
        GRPO\textdagger & \underline{0.6875} & 0.2111 & 0.6100 & \underline{0.8408} & 0.1628 & 0.2285 & 0.2179 & \underline{0.2000} & \textbf{0.3320} \\
        BPGO (only RAS) & 0.4902 & 0.1021 & 0.5580 & 0.8127 & 0.1302 & \textbf{0.2386} & \textbf{0.2306} & \textbf{0.2066} & 0.2565\\
        BPGO (only CRT) & 0.5434 & \underline{0.2485} & \underline{0.6300} & 0.8218 & \textbf{0.2278} & 0.2252 & 0.2264 & 0.1912 & 0.2900\\
        BPGO & \textbf{0.6899} & \textbf{0.2736} & \textbf{0.6500} & \textbf{0.8594} & \underline{0.2253} & \underline{0.2342} & \underline{0.2301} & 0.1964 & \underline{0.2907}\\
    \bottomrule
    \end{tabular}
    \label{tab:bgpo-vbench}
\end{table}

\paragraph{Evaluation setting.}
We evaluate BPGO on T2V (Wan2.1-1.3B) and I2V (Wan2.2-14B) tasks, both initialized from supervised fine-tuned checkpoints (Stage I output). We set group size $G=8$ and employ VideoCLIP-XL as the reward model. Our prior designs use SFT model rewards for T2V and first-frame text-alignment for I2V. We train for 200 steps on the iStock dataset (10K samples) and evaluate on 1000 test samples.

\paragraph{Metrics.}
We use VideoClipXL~\citep{wang2024videoclip} as the training reward model, measuring text-video alignment via CLIP feature similarity. For evaluation, we adopt: (1) VideoAlign~\citep{liu2025improving}, providing \textit{VideoAlign-TA} for text alignment and \textit{VideoAlign-overall} for holistic quality assessment (visual quality, motion quality, and text alignment); (2) Qwen3-VL-Embedding~\citep{qwen3vlembedding}, measuring alignment via cosine similarity in a unified multimodal representation space.

\paragraph{Results.}
Table~\ref{tab:bgpo-sota} shows BPGO consistently outperforms pretrained baselines and GRPO across all metrics, particularly on alignment-sensitive measures (VideoAlign-TA) and stability (VideoAlign-overall). On T2V, BPGO achieves 24.6\% improvement on VideoAlign-TA and improves VideoAlign-overall from -0.5411 to -0.0478. On I2V, BPGO substantially outperforms GRPO (0.6890 vs. 0.4513 on Qwen3-VL-Embedding), where GRPO's collapse indicates training instability. Human evaluation (Table~\ref{tab:bgpo-human_eval}) confirms clear preference for BPGO in overall quality and text alignment. VBench~\citep{huang2024avbench} analysis (Table~\ref{tab:bgpo-vbench}) shows BPGO achieves best performance in 6 out of 9 metrics across temporal and spatial dimensions.

% \emph{[Insert Table: main quantitative comparison across T2V/I2V/T2I.]}
% \begin{itemize}
%   \item \textbf{Source}: \texttt{BPGO/sec/4\_experiment.tex}.
%   \item \textbf{Suggested caption}: ``BPGO results across modalities.''
%   \item \textbf{Interpretation focus}: alignment-sensitive metrics (e.g., VideoAlign-TA, ImageReward) and stability (e.g., VideoAlign-overall).
% \end{itemize}

\begin{table}[!htp]
\centering
\caption{Human evaluation on 100 video pairs comparing BPGO and GRPO. BPGO shows clear human preference in both overall quality and text alignment metrics.}
\begin{tabular}{lccc}
\hline
\textbf{Metric} & \textbf{BPGO Preferred} & \textbf{Tied} & \textbf{GRPO Preferred} \\
\hline
Overall Quality & 34 & 48 & 18 \\
Text Alignment & 27 & 56 & 17 \\
\hline
\end{tabular}
\label{tab:bgpo-human_eval}
\end{table}

\begin{figure*}[!htp]
    \centering
    \includegraphics[width=\linewidth]{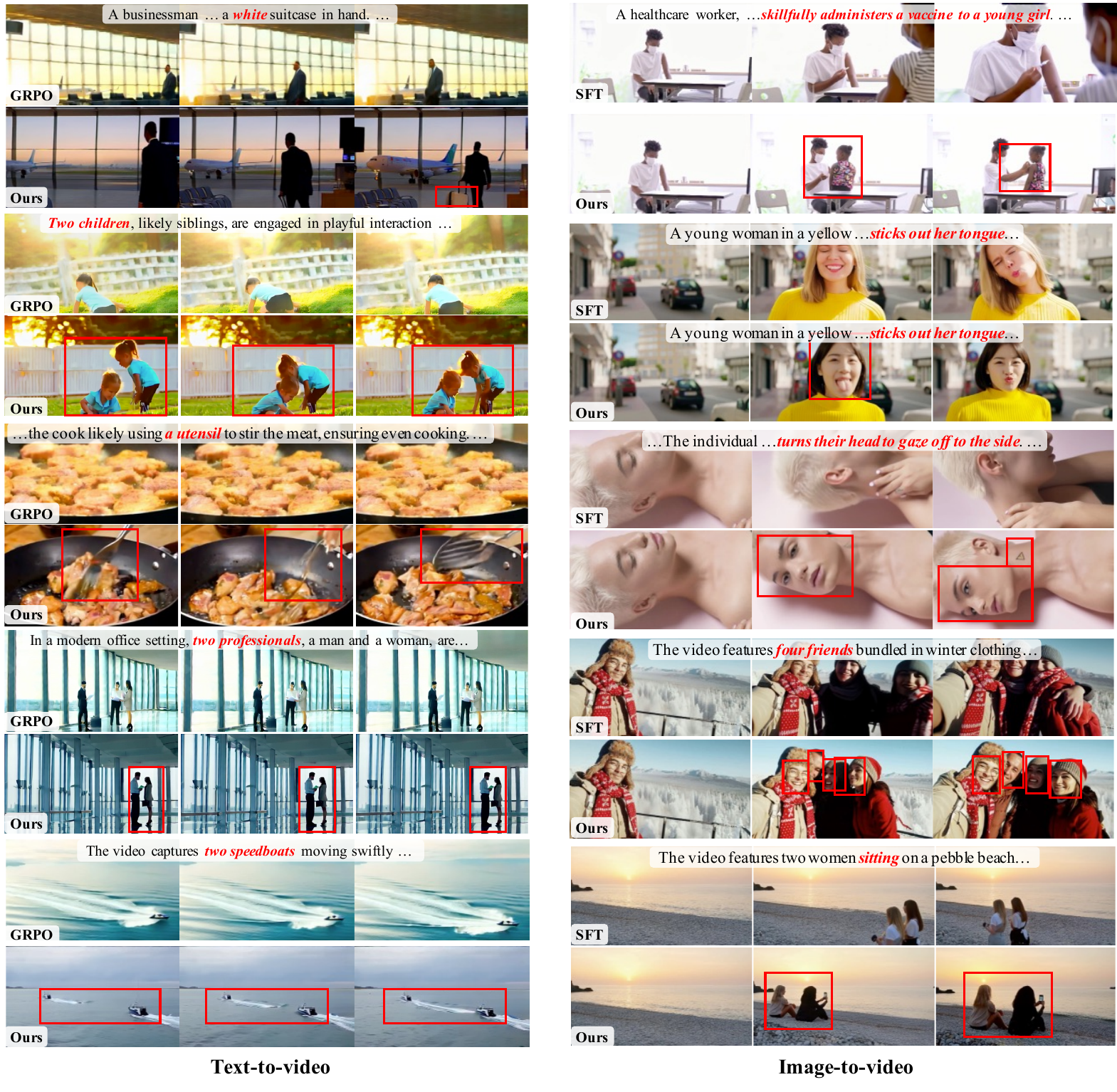}
    \caption{Qualitative comparison on text-to-video and image-to-video generation. BPGO demonstrates superior text-video alignment compared to GRPO and SFT baselines, accurately capturing fine-grained textual details (highlighted in red boxes) such as object attributes, quantities, and specific actions.}
    \label{fig:bgpo-vis}
\end{figure*}

\paragraph{Qualitative comparison.}
Figure~\ref{fig:bgpo-vis} presents qualitative comparisons on T2V and I2V generation. On T2V tasks (left), our method demonstrates superior text-video alignment compared to GRPO, accurately capturing fine-grained textual details such as object attributes (e.g., ``white suitcase''), quantities (e.g., ``two children'', ``two speedboats''), and specific actions (e.g., ``using a utensil''). The red boxes highlight regions where BGPO correctly generates semantically aligned content while GRPO produces inconsistent or missing elements. On I2V tasks (right), our method maintains better temporal consistency and identity preservation compared to the SFT baseline, generating more coherent motion while preserving subject appearance across frames. These visualizations confirm that BGPO's group-based optimization effectively improves both semantic alignment and generation quality.

\begin{figure*}[!htp]
	\centering
	\includegraphics[width=\linewidth]{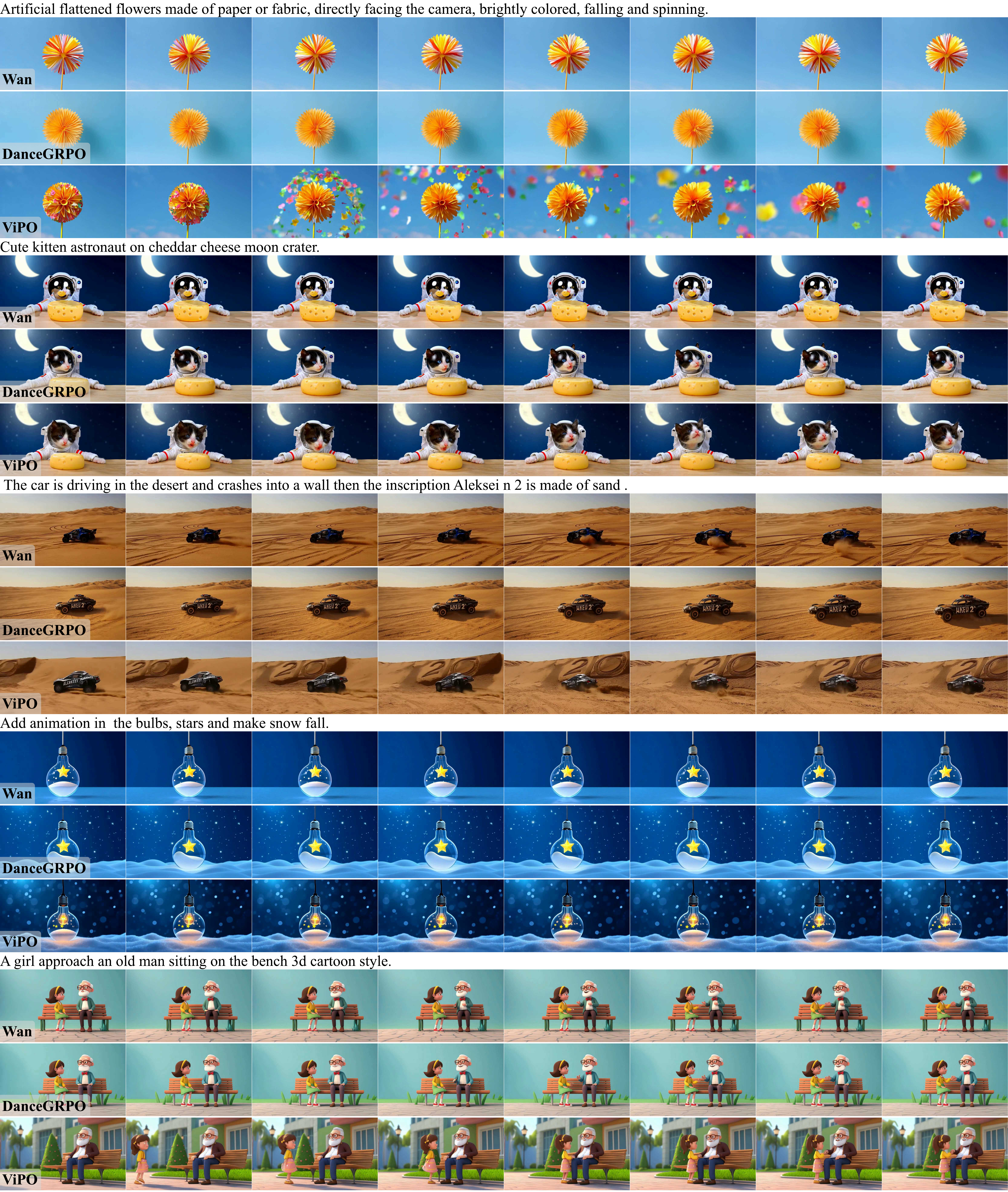}
	\caption{Qualitative case studies on Wan2.1-T2V-14B. Our approach produces more coherent motion and visually appealing frames, highlighting the effectiveness of self-paced optimization.}
	\label{fig:vipo_supp_better_one}
\end{figure*}

\subsubsection{ViPO: Structured Credit Assignment}
\label{sec:eval-vipo}

\begin{figure*}[!htp]
    \centering
    \includegraphics[width=1\linewidth]{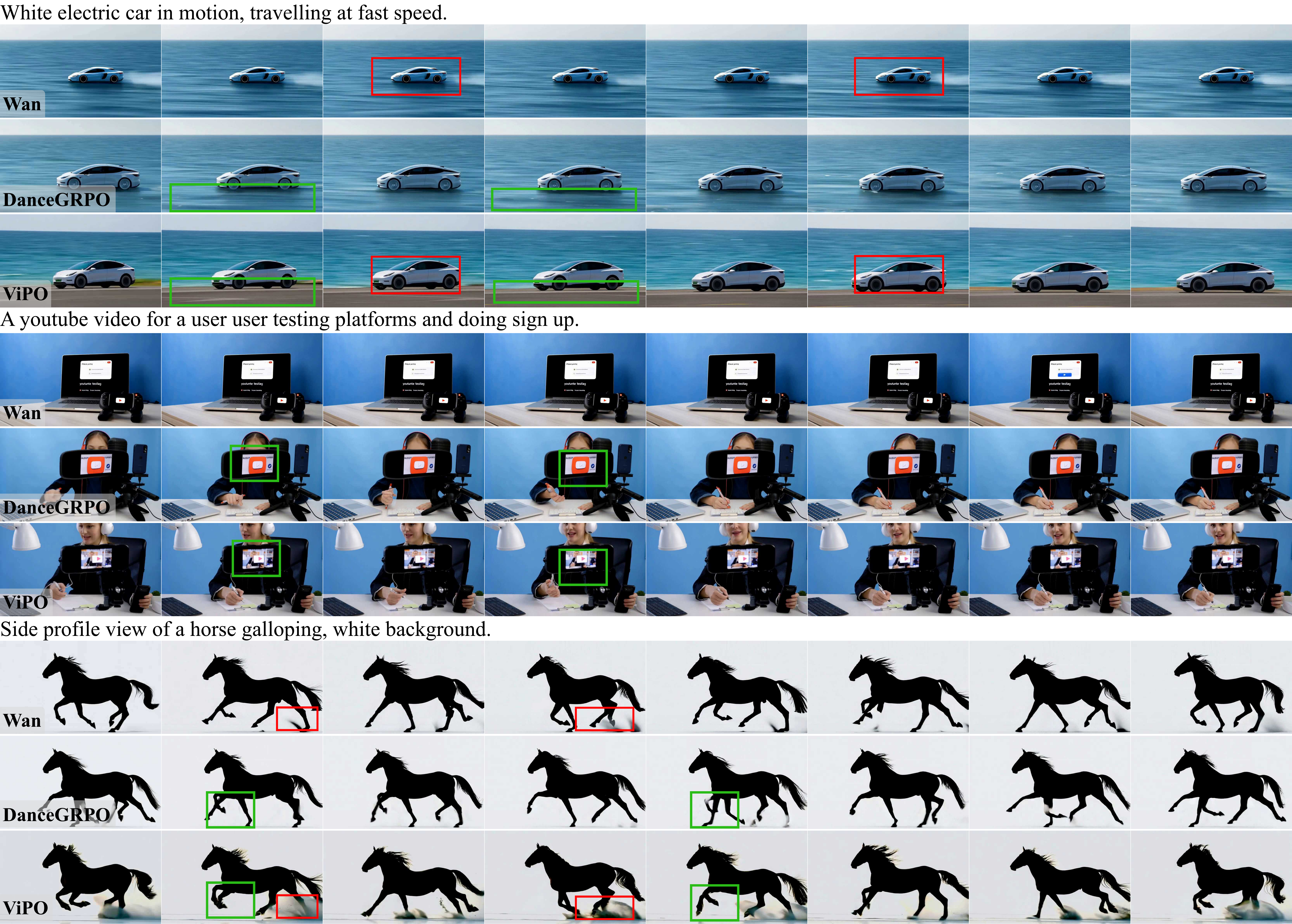} 
    \caption{Extended Qualitative comparison of VIPO with baseline methods. This figure supplements the previous comparison by providing additional representative examples, further demonstrating that VIPO consistently yields more temporally coherent motion and visually faithful frame composition, reinforcing the advantages of self-paced optimization across diverse scenarios. }
    \label{vipo_case_study}
    % \vspace{-1.5em}
\end{figure*}

\paragraph{Evaluation Setting.} 
We evaluate {VIPO} on T2V tasks using the Wan2.1-T2V-14B model. 
We set the group size to $G=32$ and employ VideoAlign~\cite{liu2025improving} as the reward model. 
All experiments are conducted on a cluster of 32 NVIDIA H100 GPUs, where each T2V task is trained for 100 steps on the 50k dataset released by DanceGRPO.

\paragraph{Results.} 
As shown in Table~\ref{tab:vipo_quant_wan}, for video generation tasks, Vipo achieves significant performance improvements across both reward model evaluation and VBench benchmarks. Notably, Vipo demonstrates a substantial gain in motion quality, improving from 0.5896 to 1.1505.More fine-grained metrics reported in Section~\ref{sec:eval-vipo} further demonstrate that our approach consistently outperforms baselines, providing deeper evidence of its effectiveness.

\begin{table}[!htp]
\caption{Quantitative comparison across detailed evaluation dimensions in VBench. ViPO consistently achieves superior performance across most dimensions.}
\centering
\small
\setlength{\tabcolsep}{4pt}
\begin{tabular}{lccc}
\toprule
\textbf{VBench} & \textbf{Wan2.1} & \textbf{DanceGRPO} & \textbf{ViPO} \\
\midrule
Dynamic Degree & 52.77 & 45.83 & \textbf{63.89} \\
Imaging Quality & 67.90 & \underline{68.31} & \textbf{68.88} \\
Multiple Objects & 69.96 & 63.18 & \textbf{74.70} \\
Color & 89.20 & 86.51 & \underline{88.97} \\
Spatial Relationship & 72.94 & 71.18 & \textbf{81.44} \\
Temporal Style & 24.12 & \underline{24.14} & \textbf{24.25} \\
Appearance Style & 21.51 & 20.91 & \underline{21.39} \\
Scene & 32.48 & 29.14 & \underline{31.90} \\
Overall Consistency & 25.06 & 25.02 & \textbf{25.32} \\
\bottomrule
\end{tabular}
\label{tab:quant_vb16_transposed}
\end{table}
\begin{table}[!htp]
\caption{Quantitative comparison results of Wan2.1. ViPO surpasses both the Wan2.1 and DanceGRPO in all out-of-domain criteria, demonstrating superior generalization.}
\centering
\small
\setlength{\tabcolsep}{2.5pt}
\begin{tabular}{lccccc}
\toprule
\multirow{2}{*}{\centering \textbf{Method}}
& \multicolumn{2}{c}{\textbf{In-domain}}
& \multicolumn{3}{c}{\textbf{Out-of-domain}} \\
\cmidrule(lr){2-3} \cmidrule(lr){4-6}
&\textbf{ VQ$\uparrow$} & \textbf{ MQ$\uparrow$} & \textbf{ Semantic$\uparrow$} & \textbf{Quality$\uparrow$} & \textbf{Total$\uparrow$}\\
\midrule
Wan2.1     &2.6219& 0.5896 &  83.36 & 71.20 & 80.92\\
DanceGRPO  & \underline{3.0935}&\underline{0.8639} & \underline{83.63}& 69.68 & 80.84  \\
ViPO &\textbf{3.5501} & \textbf{1.1515} & \textbf{83.98} & \textbf{72.59} & \textbf{81.70} \\
\bottomrule
\end{tabular}
\label{tab:vipo_quant_wan}
\end{table}

\paragraph{Case Study.}
As shown in Figure~\ref{fig:vipo_supp_better_one} and Figure~\ref{vipo_case_study}, our qualitative analysis highlights the advantages of VIPO over baseline methods. 
Compared to conventional training strategies, VIPO generates videos with smoother temporal consistency, sharper visual details, and more coherent motion trajectories. 
These improvements demonstrate that the VIPO framework successfully allocates credit to the most salient components of the video, thereby enhancing perceptual quality and yielding stronger alignment with human preferences.

\subsubsection{Self-Paced GRPO: Progressive Reward Scheduling}
\label{sec:eval-selfpaced}

We evaluate Self-Paced GRPO on VBench to assess its overall improvements in video quality. Furthermore, we conduct ablation studies disentangling progressive scheduling from direct joint training to highlight the superiority of our self-paced GRPO over strong baselines. Through comprehensive quantitative experiments, ablation analyses, and user studies, we demonstrate that Self-Paced GRPO effectively alleviates reward saturation, while substantially enhancing video quality and aligning generated content more closely with human preferences.

\begin{table*}[!htp]
\centering
\caption{Quantitative VBench results for Wan2.1‑T2V‑1.3B and Wan2.1‑T2V‑14B. 
For the 1.3B setting, we compare Wan and DanceGRPO (VideoAlign reward) with our \textbf{Self‑Paced GRPO} using Qwen2.5VL‑7B and Qwen2.5VL‑72B rewards. 
For the 14B setting, we employ Qwen2.5VL‑72B as the reward model, pretrainde Wan14B as the baseline. 
The best score for each metric is shown in \textbf{bold}.}
\label{wan vbench result}
\small
\setlength{\tabcolsep}{4pt}
\begin{tabular}{lcccc|cc}
\toprule
\multirow{2}{*}{\textbf{Metric}} 
& \multirow{2}{*}{\textbf{Wan1.3B}} 
& \multirow{2}{*}{\textbf{DanceGRPO}} 
& \textbf{Self-paced} & \textbf{Self-paced} 
& \multirow{2}{*}{\textbf{Wan14B}} 
& \textbf{Self-paced} \\
\cmidrule(lr){4-5} \cmidrule(lr){7-7}
& & & \textbf{Qwen2.5VL-7B} & \textbf{Qwen2.5VL-72B} & & \textbf{Qwen2.5VL-72B} \\
\midrule
    Aesthetic quality &60.92&58.99 & 60.97 & \textbf{62.64}& 65.33 & 60.68 \\ 
    Appearance style &20.41 &\textbf{20.70} &20.45 & 20.48& 21.35 & \textbf{21.62} \\ 
    Background consistency &96.80 &96.83 &\textbf{96.83} &96.49 & 98.36 & \textbf{98.74}\\ 
    Color &84.15&\textbf{90.05} &86.00 &84.96 & 87.98 & \textbf{89.68}   \\ 
    Dynamic degree &56.94 &58.33 &55.56 &\textbf{58.33} & 52.78 & \textbf{56.94}   \\ 
    Human action &74.00  &73.00 &\textbf{76.00} &\textbf{76.00}  & 77.00 & \textbf{80.00}   \\ 
    Image quality &67.65 &66.61  &67.63 &\textbf{68.49} & 68.03 & \textbf{68.91} \\ 
    Motion smoothness &\textbf{98.32} &98.15 &98.28 &98.23 & \textbf{98.35} & 98.30  \\ 
    Multiple objects &59.14  &57.54 &58.46  &\textbf{63.49}  & \textbf{70.27} & 69.81  \\  
    Object class &74.84&77.68 &\textbf{77.93} &74.21 & 81.72 & \textbf{82.19}   \\  
    Overall consistency  &23.63 &23.68 &\textbf{23.76}  &23.67 & 25.08 & \textbf{25.17} \\  
    Scene &22.38 & 25.79 & 20.49 &\textbf{25.94} &   \textbf{32.12} & 30.67  \\  
    Spatial relationship &68.08 &62.89 &\textbf{72.78}  &72.00 & 74.97 & \textbf{79.06}   \\ 
    Subject consistency&95.19 &\textbf{95.55} &95.10 & 95.19 & 96.49 & \textbf{96.63}  \\  
    Temporal flickering &99.38 &\textbf{99.42} &99.37 &99.35 & \textbf{99.11} & 99.01 \\  
    Temporal style &23.29 &\textbf{23.47}  &23.11  &23.15 & \textbf{24.05} & 23.98  \\  
    \addlinespace
    \textbf{Quality score} &83.15 &82.81 &83.02 &\textbf{83.53} & 84.03 & \textbf{84.59} \\  
    \textbf{Semantic score} &65.31 &66.06 &66.26  &\textbf{66.94} & 71.18 & \textbf{72.08}  \\  
    \textbf{Total score} &79.58 &79.46 &79.67 &\textbf{80.22} & 81.46 & \textbf{82.09} \\ \bottomrule 
    \end{tabular}
\end{table*}

\paragraph{Experimental Setup.}
We evaluate Self-Paced GRPO on Wan2.1-T2V tasks using both 1.3B and 14B models. 
To ensure a fair comparison, we adopt an early-stop training strategy, with all models initialized from supervised fine-tuned checkpoints. 
We set the group size to $G=8$ for the 1.3B model and $G=32$ for the 14B model. 
For reward modeling, we employ VideoAlign~\cite{liu2025improving} and Qwen2.5-VL~\cite{bai2025qwen2} as reward evaluators. 
Training is conducted for 220 steps on the 1.3B models and 100 steps on the 14B models.

\begin{table}[!htp]
\centering
\caption{Ablation study of Self-Paced GRPO on Wan2.1-T2V-1.3B. 
We compare joint training strategies with our method. 
Results are reported on VideoAlign metrics VQ, MQ, and TA, as well as the LAION aesthetic score and VBench metrics QS, SC, and OA. }
\small
\setlength{\tabcolsep}{4pt}
\begin{tabular}{lccccccc}
\toprule
\multirow{2}{*}{\textbf{Method}}
& \multicolumn{3}{c}{\textbf{VideoAlign}} 
& \textbf{LAION} 
& \multicolumn{3}{c}{\textbf{VBench}} \\
\cmidrule(lr){2-4} \cmidrule(lr){6-8}
& \textbf{VQ$\uparrow$} & \textbf{MQ$\uparrow$} & \textbf{TA$\uparrow$} 
& \textbf{Score$\uparrow$} 
& \textbf{QS$\uparrow$} & \textbf{SC$\uparrow$} & \textbf{OA$\uparrow$} \\
\midrule
Wan2.1-1.3B-T2V        & 3.448 & 0.2911 & -1.914  & 5.224 & 83.15 & 65.31 & 79.58 \\
Joint train            & 3.366 & 0.3011 & -1.077  & 5.222 & 82.69 & 66.38 & 79.80 \\
Self-paced GRPO                & \textbf{3.501} & \textbf{0.3090} & \textbf{-0.7114} & \textbf{5.252} & \textbf{83.53} & \textbf{66.94} & \textbf{80.22}  \\
\bottomrule
\end{tabular}
\label{ablation}
\end{table}

\paragraph{Results.} 
Table~\ref{wan vbench result} shows that Self-Paced GRPO achieves substantial improvements across multiple VBench dimensions. 
We attribute these gains to the self-paced GRPO framework, which adaptively tailors the optimization stage for each sample, effectively mitigating reward saturation and leading to more stable training. 
Further ablations in Table~\ref{ablation} demonstrate that the observed performance improvements stem from our self-paced algorithmic design, rather than relying on stronger reward models.

\section{Conclusion}

This report has presented a systematical post-training framework for video generation, organized as a staged optimization pipeline that progressively converts a pretrained generator into a production-oriented model. The key claim is not that any single algorithm is sufficient, but that video post-training becomes reliable only when the overall procedure is designed as a coherent stack: each stage establishes prerequisites that make the next stage both stable and meaningful under real-world constraints.

At a high level, Stage~I shapes the policy into a well-behaved, controllable interface and suppresses failure modes that would otherwise dominate downstream optimization. Stage~II performs comparative reinforcement learning under automatic feedback, with extensions that address the practical pathologies we repeatedly observe at scale: feedback that is uncertain under many-to-many correspondence, learning signals that are mismatched to localized spatiotemporal errors, and supervision that becomes non-stationary as the generator improves. Stage~III then introduces preference-based refinement to capture holistic judgments that remain difficult to express as explicit objectives, providing a complementary signal once the generator has reached a stable operating regime.

A unifying perspective across these stages is that the limiting factor in video post-training is rarely the expressiveness of the generator, but the operational quality of feedback available at training time. In practice, what determines success is whether feedback can reliably rank alternatives under the current policy distribution, whether it induces learning pressure in the right places in space and time, and whether it remains informative as competence increases. This report therefore emphasizes feedback design and its interaction with optimization as a first-order concern, and uses this lens to connect mechanisms that may otherwise appear unrelated---including reliability-aware optimization, structured credit assignment, and competence-aware supervision schedules.

Looking forward, several directions are particularly promising for pushing video post-training toward stronger robustness and broader generalization. First, feedback should become more diagnostic rather than merely scalar, exposing failure categories and uncertainty in a way that directly supports targeted optimization and debugging. Second, structured learning signals can be extended beyond perceptual allocation to incorporate richer temporal reasoning, object permanence, and viewpoint consistency, enabling better control over long-horizon degradation modes. Third, preference-based refinement can move from static pairwise comparisons toward protocols that explicitly stress temporal consistency, narrative continuity, and controllable editing, reducing presentation bias and improving alignment to real user expectations. Finally, scaling post-training will increasingly depend on system-level choices that make iteration tractable---including efficient rollout/reward serving, stable preprocessing contracts, and rigorous monitoring of drift, bias, and saturation---so that improvements accumulate predictably rather than episodically.

Overall, this report argues for treating video post-training as a disciplined optimization pipeline with explicit design principles, rather than a collection of isolated tricks. By making the stage responsibilities clear and tying algorithmic choices to the failure modes they address, the framework offers a practical foundation for developing post-training methods that are more stable, more extensible, and ultimately better aligned with deployment needs.

\section*{Contributors}
\paragraph{\textbf{Project Leaders:}} Haibin Huang, Qizhen Weng, Chi Zhang, Xuelong Li
\paragraph{\textbf{Core Contributors:}} Yuanzhi Liang, Xuan'er Wu, Yirui Liu
\paragraph{\textbf{Contributors (Listed alphabetically):}} Yijie Fang, Yizhen Fan, Ke Hao, Rui Li, Ruiying Liu, Ziqi Ni, Peng Yu, Yanbo Wang

% \section*{Acknowledgements}

% \clearpage

% \appendix
% \section{Examples}

% \clearpage

\bibliography{ref}
\bibliographystyle{authordate1}

\end{document}